\documentclass[sigconf,nonacm,balance=false,pbalance=false]{acmart}

\setcopyright{none}
\usepackage{tabularx}
\usepackage{subcaption}
\usepackage{multirow}
\usepackage{framed} 
\usepackage{pgfplots,tikz}
\pgfplotsset{compat=1.18}
\usetikzlibrary{positioning,arrows.meta}
\definecolor{darkgreen100}{RGB}{0,89,41}
\definecolor{darkgreen75}{RGB}{64,131,95}
\definecolor{darkgreen50}{RGB}{128,172,148}
\definecolor{darkgreen25}{RGB}{191,214,202}
\definecolor{darkgreen10}{RGB}{230,238,234}

\definecolor{brightgreen100}{RGB}{31,168,36}
\definecolor{brightgreen75}{RGB}{87,190,91}
\definecolor{brightgreen50}{RGB}{143,211,146}
\definecolor{brightgreen25}{RGB}{199,233,200}
\definecolor{brightgreen10}{RGB}{233,246,233}

\definecolor{raspberry100}{RGB}{130,0,78}
\definecolor{raspberry75}{RGB}{161,64,122}
\definecolor{raspberry50}{RGB}{193,128,167}
\definecolor{raspberry25}{RGB}{224,191,211}
\definecolor{raspberry10}{RGB}{243,230,237}

\definecolor{fuchsia100}{RGB}{198,0,126}
\definecolor{fuchsia75}{RGB}{212,64,158}
\definecolor{fuchsia50}{RGB}{227,128,191}
\definecolor{fuchsia25}{RGB}{241,191,223}
\definecolor{fuchsia10}{RGB}{249,230,242}

\definecolor{purple100}{RGB}{56,28,128}
\definecolor{purple75}{RGB}{106,85,160}
\definecolor{purple50}{RGB}{156,142,192}
\definecolor{purple25}{RGB}{205,198,223}
\definecolor{purple10}{RGB}{235,232,242}

\definecolor{amethyst100}{RGB}{117,59,189}
\definecolor{amethyst75}{RGB}{152,108,206}
\definecolor{amethyst50}{RGB}{186,157,222}
\definecolor{amethyst25}{RGB}{221,206,239}
\definecolor{amethyst10}{RGB}{241,235,248}

\definecolor{rust100}{RGB}{210,39,48}
\definecolor{rust75}{RGB}{221,93,100}
\definecolor{rust50}{RGB}{233,147,152}
\definecolor{rust25}{RGB}{244,201,203}
\definecolor{rust10}{RGB}{251,233,234}

\definecolor{tangerine100}{RGB}{237,112,0}
\definecolor{tangerine75}{RGB}{242,148,64}
\definecolor{tangerine50}{RGB}{246,184,128}
\definecolor{tangerine25}{RGB}{251,219,191}
\definecolor{tangerine10}{RGB}{253,241,230}

\definecolor{cobalt100}{RGB}{0,50,160}
\definecolor{cobalt75}{RGB}{64,101,184}
\definecolor{cobalt50}{RGB}{128,153,208}
\definecolor{cobalt25}{RGB}{191,204,231}
\definecolor{cobalt10}{RGB}{230,235,246}

\definecolor{azure100}{RGB}{0,113,222}
\definecolor{azure75}{RGB}{64,148,230}
\definecolor{azure50}{RGB}{128,184,239}
\definecolor{azure25}{RGB}{191,220,247}
\definecolor{azure10}{RGB}{230,241,252}

\floatstyle{ruled}
\newfloat{algorithm}{tbp}{loa}
\floatname{algorithm}{Algorithm}
\floatstyle{plain}

\title[DiRAFT: Privacy-Aware Financial Transcript Synthesis]{DiRAFT: Privacy-Aware Financial Transcript Synthesis from Aggregate Discourse Priors}
\titlenote{The paper is accepted at the Financial AI Security, Privacy and Safety Workshop at ICAIF 2026 (non-archival).}

\author{Yunbai Zhang}
\email{yunbai\_zhang@vanguard.com}
\affiliation{%
  \institution{The Vanguard Group}
  \city{Malvern}
  \state{Pennsylvania}
  \country{USA}
}

\author{Yunyi Gao}
\email{yunyi\_gao@vanguard.com}
\affiliation{%
  \institution{The Vanguard Group}
  \city{Malvern}
  \state{Pennsylvania}
  \country{USA}
}

\author{Han-Tai Shiao}
\authornote{Corresponding author. \copyright~2026 The Vanguard Group, Inc. All rights reserved.}
\email{han-tai\_shiao@vanguard.com}
\affiliation{%
  \institution{The Vanguard Group}
  \city{Malvern}
  \state{Pennsylvania}
  \country{USA}
}

\makeatletter
\patchcmd{\@mktitle@iii}{\par\bigskip}{\par\smallskip}
  {}{\errmessage{Could not adjust title spacing}}
\patchcmd{\@mkauthors@iii}{\par\bigskip}{\par\smallskip}
  {}{\errmessage{Could not adjust author spacing}}
\makeatother

\begin{document}

\begin{abstract}
    Using customer conversations to train and evaluate financial language models raises
    confidentiality concerns because non-public financial facts can remain after direct
    identifiers are masked. Synthetic transcripts offer an alternative data source, but their
    discourse fidelity and residual confidentiality risks require assessment. We introduce
    DiRAFT (Discourse Realization from Aggregate First-order Transitions), which extracts
    financial concept graphs and records the order in which concepts are first introduced. It
    pools adjacent concept transitions across training records and samples new graph plans
    from aggregate transition, starting-concept, and plan-length priors. A discourse layer
    guides text realization without supplying an individual source passage or graph. We assess
    privacy-related risks through two complementary evaluations: corpus-level structural
    overlap and source-grounded checks for specific non-public investor facts. The structural
    assessment covers two financial corpora and three generation backends. The
    confidential-fact assessment uses human review of 1,400 investor-call outputs across seven
    strategies and two backends. We also evaluate transition fidelity, content recoverability,
    and trajectory diversity. Stochastic sampling provides a favorable balance between
    transition fidelity and structural overlap, with consistent advantages over greedy and
    top-3 on investor calls. The confidential-fact assessment likewise detects fewer matches
    for stochastic sampling than these alternatives on both backends covered by the fact
    assessment. These complementary assessments provide empirical evidence on residual
    confidentiality risks without establishing a formal guarantee.
\end{abstract}

\maketitle

\section{Introduction}

    Customer conversations provide a potential source of training and evaluation
    data for financial language models. However, these records can contain
    non-public portfolio positions, transaction intentions, and payment
    obligations even after direct identifiers are masked. Synthetic
    transcripts\footnote{We use \emph{transcript} for the spoken-record genre and
    \emph{document} as the formal term throughout the technical development;
    both denote the same artifact.}
    offer a potential alternative data source, creating a need to assess both
    their discourse fidelity and their reproduction of confidential source
    information.

    A source graph used to condition generation can retain distinctive content
    and structure even when the original wording is absent. This motivates
    generating from patterns pooled across records while examining what
    source-specific information remains in the outputs. Graph representations
    make selected financial content explicit, providing concepts that can be
    pooled across records and arranged into new content plans. Their order also
    matters: concept co-occurrence indicates which items appear together, but
    not which tends to be introduced next. Here, we model transitions between
    financial concepts in their order of first introduction.

    \subsection{Related Work}

    We situate our contribution relative to three lines of prior research: graph-guided
    synthetic generation; knowledge paths and discourse flow; and privacy of synthetic text.

    \begingroup
    \setlength{\parskip}{0pt}

    \textbf{Graph-guided synthetic generation.}
    Knowledge graphs organize entities and relations~\citep{hogan2021knowledge}, and
    graph-to-text models use this structure for text
    realization~\citep{koncel2019text,ribeiro2020modeling}. Synthetic-data methods
    assign graphs distinct generative roles. EntiGraph generates relation-focused text
    from within-document entity subsets for continued
    pretraining~\citep{yang2025synthetic}. SoG traverses an entity co-occurrence graph
    to retrieve cross-document passages, emphasizing multi-hop and long-tail
    knowledge~\citep{ma2025synthesize}. GraphSynth samples valid attribute
    configurations from a probabilistic factor graph under schema- and
    ontology-constrained decoding~\citep{cheng2026graphsynth}. These methods chiefly
    decide which facts, passages, or compatible attributes to combine; they do not
    estimate how a new financial document should unfold.

    \textbf{Knowledge paths and discourse flow.}
    Other work models discourse through entity transitions, knowledge-graph walks,
    and discourse concept flows~\citep{barzilay2008modeling,moon2019opendialkg,li2023enhancing}, as well
    as corpus-learned domain content order and adjacent-turn transition
    graphs~\citep{barzilay2004catching,xu2020conversational}. MP2D, for
    example, constructs topic-shifting conversational question answering through
    directed walks over knowledge-graph triples and retrieves passages for visited
    entities, making successive topics relationally
    interpretable~\citep{hwang2024mp2d}. These knowledge-graph paths inherit candidate
    directions from graph semantics and traversal, whereas corpus-learned models
    operate at latent-topic or adjacent-turn level. We instead estimate concepts' first-introduction order in human-authored
    financial documents. Graph connectivity explains why topics may connect; our transition prior
    captures how they typically unfold.

    \textbf{Privacy of synthetic text.}
    Models can reveal training-set membership or memorized records, and synthetic
    text is not automatically
    private~\citep{shokri2017membership,carlini2021extracting,carlini2023quantifying,nasr2025scalable,meeus2025canary}.
    Text anonymization also considers indirect identifiers and confidential
    attributes~\citep{pilan2022tab}. Privacy-aware synthesis includes differentially
    private (DP) fine-tuning, including generators guided by DP topic
    histograms~\citep{yue2023synthetic,tan2025ctcl}, and private evolution or
    prediction~\citep{xie2024differentially,amin2024private}. Entity-aware controlled
    generation offers another approach to limiting identifier
    disclosure~\citep{zhao2025controlled}. Our study assesses corpus-level structural
    overlap and matches to specific non-public investor facts alongside discourse
    fidelity. These assessments provide empirical evidence rather than a formal
    privacy guarantee.
    \par
    \endgroup

    \subsection{Our Approach and Contributions}

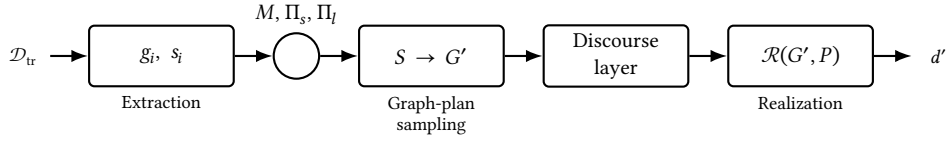
\begin{figure*}[t]
    \centering
    \begin{tikzpicture}[
        >={Latex[length=2.2mm]},
        font=\small,
        proc/.style={
            draw, thick, black, rectangle, rounded corners=2pt, align=center,
            font=\small, inner sep=3pt,
            text width=17mm, minimum height=8mm
        },
        op/.style={draw, thick, black, circle, inner sep=0pt, minimum size=6mm},
        docnode/.style={align=center, text width=5mm, font=\footnotesize},
        caption/.style={align=center, text width=17mm, font=\footnotesize},
        arrow/.style={->, thick, black},
    ]
        \node[docnode]                (s1) {$\mathcal{D}_{\mathrm{tr}}$};
        \node[proc, right=5mm of s1]  (s2) {$g_i,\ s_i$};
        \node[op, right=5mm of s2]  (s3) {};
        \node[proc, right=5mm of s3]  (s4) {$S \rightarrow G'$};
        \node[proc, right=5mm of s4]  (s5) {Discourse\\ layer};
        \node[proc, right=5mm of s5]  (s6) {$\mathcal{R}(G',P)$};
        \node[docnode, right=5mm of s6]  (s7) {$d'$};

        \node[above=0mm of s3] {$M$, $\Pi_s$, $\Pi_l$};
        \node[caption, below=0mm of s2] {Extraction};
        \node[caption, below=0mm of s4] {Graph-plan\\ sampling};
        \node[caption, below=0mm of s6] {Realization};

        \draw[arrow] (s1) -- (s2);
        \draw[arrow] (s2) -- (s3);
        \draw[arrow] (s3) -- (s4);
        \draw[arrow] (s4) -- (s5);
        \draw[arrow] (s5) -- (s6);
        \draw[arrow] (s6) -- (s7);
    \end{tikzpicture}
    \caption{DiRAFT's graph-mediated synthesis: extraction, aggregate-prior plan sampling, and discourse-guided realization.}
    \Description{A left-to-right pipeline extracts graphs and ordered concept sequences from a training corpus,
    estimates aggregate transition, start-concept, and plan-length priors,
    samples a new graph plan, applies a discourse layer, and realizes a synthetic document.}
    \label{fig:workflow}
\end{figure*}

    We propose DiRAFT (Discourse Realization from Aggregate First-order Transitions), an aggregate transition-prior framework that learns corpus-level
    priors over concept transitions, starting concepts, and plan lengths. The
    framework samples new graph plans under alternative transition policies and
    realizes them as text through a lightweight, controllable discourse layer, as summarized in
    Figure~\ref{fig:workflow}. Because transitions are pooled across records, the
    prior captures corpus-level discourse patterns while keeping individual source
    records outside the realization interface.
    Stochastic sampling then balances fidelity to these learned discourse flows
    against two risks: reproducing the structure of individual source records and
    repeatedly following a small set of high-frequency paths.

    Our contributions are threefold. \textbf{(1) Privacy-aware synthesis.} We introduce
    DiRAFT, combining aggregate discourse priors, stochastic sampling, and controllable
    discourse realization without conditioning the realizer on an individual source record.
    \textbf{(2) Privacy assessment.} We evaluate residual risks through corpus-level
    structural overlap and source-grounded checks for specific non-public investor facts.
    \textbf{(3) Generation quality.} We assess transition fidelity, round-trip content
    recoverability, and trajectory diversity.

\section{Methodology}\label{sec:methodology}
\subsection{Task and Representation}
DiRAFT aims to preserve corpus-level discourse patterns while limiting source-specific structural reproduction. It estimates aggregate priors from training documents, samples a content plan, and realizes it with a discourse plan. Figure~\ref{fig:workflow} shows the pipeline.

Let $\mathcal{D}=\{d_i\}_{i=1}^{n}$ be a corpus of financial documents, with training partition $\mathcal{D}_{\mathrm{tr}}$. Pipeline $\mathcal{E}$ combines LLM concept extraction with deterministic ordering to obtain a directed graph $g_i$ and sequence $s_i=(z_{i,1},\ldots,z_{i,l_i})$. Here, $l_i$ is sequence length and $z_{i,t}$ the $t$-th concept. Each canonical concept is retained once, ordered by first-mention turn and within-turn text position. Edges link consecutive concepts in introduction order. Training documents define generation vocabulary $V$: financial topics, entities, products, and events.

Generation uses sampled plans rather than individual source passages or graphs. Extraction and prior estimation require source records; aggregation gives no formal privacy guarantee.

\subsection{Aggregate Priors and Plan Sampling}
From training sequences, we estimate the start-concept distribution $\Pi_s$, sequence-length distribution $\Pi_l$, and directed transition matrix $M$. The first two record empirical frequencies of initial concepts and sequence lengths. For concepts $u,v\in V$, adjacent-transition counts define
\begin{equation}
\begin{split}
T_{uv}&=\sum_{i:d_i\in\mathcal{D}_{\mathrm{tr}}}\sum_{t=1}^{l_i-1}\mathbf{1}\{z_{i,t}=u,\ z_{i,t+1}=v\},\\
M_{uv}&=T_{uv}\big/\sum_{w\in V}T_{uw}.
\end{split}
\end{equation}
Here, $\mathbf{1}$ is the indicator function. For each nonzero row, $M_{uv}$ gives the probability of introducing $v$ after $u$. Rows are normalized without smoothing, leaving unobserved transitions at zero probability.

Stochastic sampling draws $z_1\sim\Pi_s$ and a target length $l\sim\Pi_l$, then repeatedly samples $z_{t+1}\sim M_{z_t,\cdot}$. Sampling stops at length $l$ or when no positive-probability successor exists. Unlike extracted first-introduction sequences, the sampled sequence (content plan $S$) permits repeats and cycles. Graph plan $G'$ represents $S$ with directed edges between consecutive concepts.

Four comparison strategies share the same start and length priors. Greedy selects the most probable successor; top-3 samples from the renormalized probabilities of up to three leading successors. Co-occurrence sampling uses row-normalized symmetric document-level co-occurrence counts, discarding observed discourse order. Uniform-random samples remaining concepts independently and uniformly from $V$ with replacement. All walk-based strategies use the same stopping rule. Separately, source-graph conditioning supplies one randomly selected training graph as a structural exemplar, without source text.

\subsection{Discourse Planning and Realization}
The content plan specifies what to discuss and in what order; the discourse plan $P$ specifies how to organize those concepts into dialogue. Using aggregate script priors $\Psi$ estimated from training documents, the discourse layer samples call goal, tone, speaker pattern, and turn budget. It distributes concepts across body turns, reserves opening and closing turns, and checks mandatory-concept coverage. Speaker count is configurable (default: two). Table~\ref{tab:discourse-controls} in Appendix~\ref{app:extended-methods} lists the control definitions and settings.

DiRAFT's LLM realizer produces $d'=\mathcal{R}(G',P)$ from the sampled graph and discourse plans, without an individual source transcript or graph. The transcript is re-extracted by $\mathcal{E}$ for evaluation. Algorithm~\ref{alg:diraft-synthesis} in Appendix~\ref{app:extended-methods} gives pseudocode; the appendix also details transition policies and discourse-control rules. Appendix~\ref{app:prompts} provides prompt templates.

\section{Evaluation and Results}\label{sec:evaluation}
We assess privacy, transition fidelity, human transcript preferences, sensitivity to concept order, content preservation, and trajectory diversity without combining them into a scalar score.
\subsection{Experimental Setup}
We use 1,357 proprietary investor calls from 846 investors within one month, with personally identifiable information masked and investor-disjoint training/held-out partitions. Confidentiality prevents public release. Calls are modeled separately, without longitudinal relationships. We also use 297 public S\&P~500 earnings calls from February 2025 with a document-level 80/20 split~\citep{kurry2025sp500earnings}.

Except for constructed target sequences in Appendix~\ref{app:single-call-effects}, all corpus-derived generation inputs come exclusively from training data. Opus~4.6 is the fixed extractor; generators are Opus~4.6, GPT-5.4-mini, and Haiku~4.5 (Opus, GPT, and Haiku). Except for the coherence generator--judge comparisons in Appendix~\ref{app:control-evaluation}, all LLM-based judging used Gemini~3.1 Pro, which was not used for transcript generation. The separate judge family was chosen to reduce potential self-preference bias. Structural-overlap evaluation includes 1,000 eligible recovered graphs per setting. Appendices~\ref{app:full-setup} and~\ref{app:prompts} detail partitions, references, shared plans and seeds, filtering, and investor-call prompts.

\textbf{Extractor validation.}
To assess method-dependent extraction errors in structural evaluation, we compared Opus extraction with human annotations of 60 held-out real and 540 synthetic investor calls. Macro-averaged concept precision and recall were 92.4--95.6\% and 91.4--95.5\%; corresponding transition scores were 88.2--91.9\% and 87.6--91.7\%. These results support overall extraction accuracy on investor-call texts (Appendix~\ref{app:extraction-validation}).

\subsection{Privacy}\label{sec:eval-privacy}

Synthetic generation does not by itself resolve confidentiality concerns,
because outputs may still reproduce source-specific structure or non-public
investor information. We conduct two complementary assessments:
corpus-level structural overlap and sample-based confidential-fact detection
through human review. The former measures resemblance to individual real
records, while the latter checks for specific non-public investor facts
supported by source texts. A controlled in/out experiment additionally
measures the structural effect of an added call-level concept sequence.

Table~\ref{tab:privacy-fidelity-summary} summarizes investor-call privacy and
transition fidelity for Opus~4.6 and GPT-5.4-mini. Full results and procedures,
including Haiku~4.5 structural-overlap and transition-fidelity results, appear in
Appendices~\ref{app:confidentiality-audit}, \ref{app:structural-overlap},
and~\ref{app:supporting-evaluation}.

\textbf{Structural Overlap.}
Structural overlap measures how closely a synthetic output resembles an
individual real record in its concepts and their ordering. This corpus-level
screen checks for source-specific structure that may remain even when
generation uses aggregate priors. Using Opus~4.6 as the fixed extractor, we compare each
recovered graph against all training and held-out real graphs, selecting the
reference with maximum entity Jaccard for all three metrics. Entity Jaccard
measures unordered concept-set overlap; directed-transition Jaccard measures
overlap in adjacent-concept pairs; longest common transition subsequence length
(LCT) counts transition pairs shared in order, allowing gaps. We report their
means and the upper-tail flagged-overlap rate, the proportion with entity Jaccard
$\geq0.4$ (Appendix~\ref{app:structural-overlap}).

Across both datasets, stochastic sampling had lower values on all four measures than greedy in all six settings and top-3 in five, including all three investor-call backends. Its investor-call flagged rates were 2.1--7.2\%, versus 20.3--53.2\% for greedy, 13.1--38.4\% for top-3, and 14.3--18.5\% for source-graph conditioning (Table~\ref{tab:structural-overlap-results}). The exception was S\&P--Opus: top-3 had lower entity and transition Jaccard and flagged overlap, whereas stochastic had lower LCT. Stochastic and co-occurrence offered lower-overlap alternatives to greedy; uniform-random generally had the least overlap. These scores establish neither confidentiality violations nor realistic discourse flow (Section~\ref{sec:eval-transition-fidelity}).

\textbf{Confidential-Fact Detection.}
Low structural overlap does not exclude reproduction of a specific non-public
fact, such as an investor's planned fund sale. We therefore review outputs against source texts
for confidential-fact matches, including details absent from recovered graphs.

Across seven strategies and two backends (Opus~4.6 and GPT-5.4-mini), we audited
100 investor-call outputs per cell (1,400 total).
Two reviewers independently searched the full source corpus, blinded to strategy
and backend; a third adjudicated disagreements and uncertain judgments. Confirmed
matches required paired source/output spans expressing the same specific
non-public investor fact; generic topics or public product names alone did not
qualify. Rates count each output with a confirmed match once, retaining uncertain cases
in the denominator
(Appendix~\ref{app:confidentiality-audit}).

Stochastic yielded 2.0\% detected matches per backend (2/100 each; 95\% Wilson CI, 0.6--7.0\%), below source-graph conditioning (20.0--23.0\%), greedy (16.0--17.0\%), and top-3 (9.0--13.0\%). Uniform-random and co-occurrence yielded 0--3.0\% (Table~\ref{tab:privacy-fidelity-summary}). Scenario-only prompting yielded 1.0\% on both backends (Appendix~\ref{app:confidentiality-audit}).

\begin{table*}[t]
    \caption{Privacy--fidelity results for six corpus-conditioned investor-call strategies. Privacy indicators are flagged structural overlap (nearest-reference entity Jaccard $\geq0.4$; 1,000 eligible outputs) and detected confidential-fact matches (100 audited outputs per cell; 95\% Wilson CI; uncertain cases retained). Fidelity is assessed by weighted transition Jensen--Shannon divergence (JSD) between synthetic and held-out real next-concept distributions on shared conditioning concepts. Columns report separate evaluations; lower is better within each measure, and bold marks backend-specific minima.}
    \label{tab:privacy-fidelity-summary}
    \centering
    \footnotesize
    \setlength{\tabcolsep}{4pt}
    \renewcommand{\arraystretch}{1.05}
    \begin{tabular*}{\textwidth}{@{\extracolsep{\fill}}lrrrrrr@{}}
        \toprule
        & \multicolumn{4}{c}{Privacy} & \multicolumn{2}{c}{Transition fidelity} \\
        \cmidrule(lr){2-5}\cmidrule(l){6-7}
        & \multicolumn{2}{c}{Flagged structural overlap (\%)} & \multicolumn{2}{c}{Detected confidential-fact matches (\%)} & \multicolumn{2}{c}{Weighted transition JSD} \\
        \cmidrule(lr){2-3}\cmidrule(lr){4-5}\cmidrule(l){6-7}
        Strategy & Opus~4.6 & GPT-5.4-mini & Opus~4.6 & GPT-5.4-mini & Opus~4.6 & GPT-5.4-mini \\
        \midrule
        \textbf{Stochastic} & 2.1 & 7.2 & 2.0 (0.6--7.0) & 2.0 (0.6--7.0) & \textbf{0.563} & \textbf{0.478} \\
        Greedy & 20.3 & 53.2 & 16.0 (10.1--24.4) & 17.0 (10.9--25.5) & 0.710 & 0.722 \\
        Top-3 & 13.1 & 38.4 & 9.0 (4.8--16.2) & 13.0 (7.8--21.0) & 0.649 & 0.621 \\
        Uniform-random & \textbf{0.6} & \textbf{0.4} & \textbf{1.0} (0.2--5.4) & \textbf{0.0} (0.0--3.7) & 0.931 & 0.988 \\
        Co-occurrence & 3.3 & 4.9 & 2.0 (0.6--7.0) & 3.0 (1.0--8.5) & 0.806 & 0.827 \\
        Source-graph conditioning & 14.3 & 18.5 & 20.0 (13.3--28.9) & 23.0 (15.8--32.2) & 0.767 & 0.771 \\
        \bottomrule
    \end{tabular*}
\end{table*}

\textbf{Single-Call Effects.}
The preceding assessments examine source information in outputs, but not how adding one call changes them. We therefore tested 30 constructed financial-concept sequences in a controlled in/out experiment. Each represented an additional call whose concepts were present but adjacent transitions absent in a background of 100 real investor calls. Each sequence was separately tested for reproduction before and after addition under identical generation settings. Mean full-path reproduction after re-extraction increased from 0.02\% to 0.12\% for DiRAFT and from 0.02\% to 0.83\% for source-path conditioning. DiRAFT also showed a smaller planning-stage increase, supporting lower observed influence of an individual added sequence on structural reproduction (Appendix~\ref{app:single-call-effects}).

\subsection{Discourse Fidelity}\label{sec:eval-transition-fidelity}

Privacy measures alone do not establish that synthetic transcripts are realistic:
low structural overlap and few detected confidential-fact matches can coexist with
implausible discourse, such as abrupt shifts between unrelated financial topics.
We therefore assess corpus-level transition fidelity and transcript-level human preference; order ablation tests the role of sampled concept order.

\textbf{Transition Fidelity.}
Transition fidelity assesses whether synthetic transcripts preserve corpus-level
patterns of which concept is introduced next. We extract first-introduction concept
sequences and compare adjacent-transition distributions against held-out real
transcripts using weighted Jensen--Shannon divergence (JSD); lower values indicate
closer agreement. Comparisons cover conditioning concepts shared by both collections
and are weighted by held-out frequency, assessing conditional transition similarity,
not concept coverage. Training data supply generation priors; held-out data provide
the fidelity reference (Appendix~\ref{sec:eval-realism}).

Stochastic sampling achieved the lowest JSD on investor calls across all three
backends (0.478--0.563) and on S\&P~500 calls with Opus and GPT; top-3 narrowly led
with Haiku (0.726 versus 0.727; Table~\ref{tab:transition-fidelity-results}). It
had 0.161--0.334 lower JSD than scenario-only prompting across all six
dataset--backend settings, supporting closer alignment with held-out transition patterns. On both audited investor-call backends, stochastic's lowest JSD and low
flagged-overlap and detected-match rates support a favorable privacy--fidelity balance. Uniform-random, co-occurrence, and scenario-only prompting also had low
detected-match rates but weaker transition fidelity.

\textbf{Generated Transcript Preference.}
Aggregate transition similarity does not establish whether a complete dialogue has clear topic connections, motivated shifts, or a plausible financial discussion. We therefore compare human preferences for stochastic versus scenario-only prompting, source-graph-conditioned, and co-occurrence outputs. Two annotators each judged the same 100 Opus~4.6 investor-call pairs per comparison (200 judgments). Tie-adjusted preference is the fraction favoring stochastic plus half the tied fraction; 50\% indicates no aggregate preference. Pair-bootstrap confidence intervals retain both annotators' judgments, with Bonferroni adjustment across the three comparisons (Appendix~\ref{app:strategy-preference-evaluation}).

Tie-adjusted preferences for stochastic were 75.25\%, 72.50\%, and 62.75\%, respectively; all adjusted confidence intervals exceeded 50\%. Inter-annotator agreement was 55--59\%. Complementing the distributional evidence, human judgments favored stochastic outputs as complete transcripts in this Opus setting.

\textbf{Order ablation.}
To examine whether sampled concept order matters to perceived topic progression,
two annotators independently judged 100 investor-call pairs per backend comparing
outputs from original stochastic plans and order-shuffled counterparts. Tie-adjusted preferences
for Original were 64.75\% (Opus) and 61.75\% (GPT), with both 95\% pair-bootstrap CIs
above 50\% (Table~\ref{tab:order-preference}; Appendix~\ref{app:order-evaluation}),
supporting the relevance of sampled order to perceived topic progression in these
comparisons.

\subsection{Content Preservation and Diversity}\label{sec:eval-content-diversity}
\textbf{Content preservation.}
A sampled plan is useful only if its intended concepts and connections survive realization.
We compare plans with re-extracted graphs after node alignment using
grounding-adjusted node and directed-edge macro-F1 across documents. Recall measures recovery of planned items; precision also accepts document-supported enrichment
(Appendix~\ref{sec:eval-recoverability}). Source-graph conditioning lacks paired
plans and is excluded. Across both corpora and all backends, most scores were at least 0.90 (minimum 0.886; Table~\ref{tab:recoverability-results}). Planned content thus generally remains recoverable after realization. High scores for greedy and uniform-random reflect plan execution rather than overall generator quality.

\textbf{Trajectory diversity.}
Plausible transitions can form repetitive paths, so we assess collapse using the two-step unique path ratio (UPR-2). Lower UPR-2 indicates greater repetition; shared plans and backend-specific recoveries are evaluated against full-corpus real graphs (Appendix~\ref{sec:eval-diversity}). On investor calls, stochastic scored 0.937 on plans (greedy, 0.284; top-3, 0.467) and 0.964--0.971 on recoveries (real reference $\approx 0.99$; Table~\ref{tab:diversity-results}), retaining varied paths alongside transition fidelity. Uniform-random and co-occurrence also showed high diversity but weaker fidelity, reinforcing diversity as a collapse check rather than an objective to maximize.

\section{Discussion and Limitations}\label{sec:discussion}
On investor calls, stochastic combined lower transition divergence and fewer detected confidential-fact matches than greedy, top-3, and source-graph conditioning on both audited backends.

Adding the discourse layer reduced model-judged mean coherence by 0.12--0.73 points (1--5 scale), indicating a controllability--coherence trade-off (Appendix~\ref{app:control-evaluation}).

No formal or investor-level privacy guarantee follows. Manual search may miss matches; detected rates establish neither exhaustive disclosure prevalence nor causal leakage, especially with held-out references. One extractor standardizes comparisons, but sensitivity to extractor choice is untested; abstraction may miss lexical leakage or memorized spans. Structural results cover two corpora and three backends; investor-call confidentiality and order studies cover two backends, and strategy preferences only Opus~4.6. Generalization to other domains or weaker models remains open. Human transcript preferences are also conditional on the observed two-annotator panel and evaluated settings. Finally, the reported evaluations assess transcript properties but do not establish downstream utility for training or evaluating financial language models. Future work includes controlled memorization tests, broader discourse evaluation, typed-relation generation, and formal privacy mechanisms.

\clearpage
\bibliographystyle{ACM-Reference-Format}
\bibliography{synthetic}

\clearpage
\appendix
\setcounter{topnumber}{3}
\setcounter{dbltopnumber}{3}
\setcounter{totalnumber}{5}
\renewcommand{\topfraction}{0.9}
\renewcommand{\dbltopfraction}{0.85}
\renewcommand{\textfraction}{0.08}
\renewcommand{\floatpagefraction}{0.75}
\renewcommand{\dblfloatpagefraction}{0.75}
\setlength{\textfloatsep}{10pt plus 2pt minus 2pt}
\setlength{\dbltextfloatsep}{10pt plus 2pt minus 2pt}
\setlength{\floatsep}{8pt plus 2pt minus 2pt}
\setlength{\dblfloatsep}{8pt plus 2pt minus 2pt}
\clubpenalty=10000
\widowpenalty=10000
\section*{Appendix}

\section{Confidential-Fact Detection: Review Protocol}\label{app:confidentiality-audit}

    This assessment checks whether synthetic investor calls reproduce specific
    non-public facts supported by source records. It complements structural
    screening by examining semantic content in the text, including facts omitted
    by graph extraction.

\subsection{Scope and review materials}

    The audit covered 100 outputs from each of seven strategies with Opus~4.6 and
    GPT-5.4-mini: 1,400 distinct outputs, each independently reviewed by two annotators.
    The counting unit was the generated output, not a matched fact or reviewer
    judgment. All sampled outputs completed the prescribed review procedure.
    The small-graph filter for concept Jaccard did not exclude outputs from this
    text-based audit. Reviewers searched the same collection of 1,357 masked source
    calls with existing full-text search, blinded to strategy and backend.
    Source-graph conditioning received no additional source-exemplar cues.
    Any recovered graph served only as a navigation aid; eligibility and evidence
    depended on the text, including facts absent from the recovered graph.
    Call goal, tone, speaker pattern, and turn budget were fixed across the six
    corpus-conditioned strategies and both backends, including source-graph conditioning,
    while strategy-specific content varied. Scenario-only prompting used the generic
    financial-call template in Appendix~\ref{app:prompt-scenario}, without source
    records, extracted graphs, concept inventories, or corpus-estimated priors.
    Its outputs were assessed under the same confidentiality criteria and search
    procedure. The comparison evaluates the complete generation procedures rather
    than isolating the effect of their conditioning inputs.


\subsection{Confidentiality criteria}\label{app:audit-criteria}

    Confidential facts were non-public, client-specific financial positions,
    circumstances, transaction instructions, or individualized financial arrangements.
    Reviewers separately assessed semantic agreement, confidentiality, and the
    specificity of the shared information. A fact comprised a subject or role,
    relation, object or value, and material conditions. Public information did not
    qualify merely because it appeared in a private call. Public product names,
    generic macroeconomic topics, and general investment advice alone were insufficient.

    Semantic agreement allowed paraphrase and verified aliases, but required
    consistency in the relation and material attributes, including transaction
    direction, amount, timing, and conditions when present. A contradictory attribute
    prevented confirmation of that complete fact; an independently supported component,
    such as a specific holding relation, could still qualify. One concrete confidential
    fact was sufficient; the entire distinctive combination need not be reproduced.
    Shared nouns without a matching relation were insufficient. Limited search did
    not establish corpus uniqueness or investor identifiability.

    Financial expertise informed the taxonomy and search guidance.
    Table~\ref{tab:audit-taxonomy} illustrates categories and anchor types rather than
    the complete internal taxonomy. Anchors guide discovery, not the final judgment.
    Ambiguity about semantic agreement or confidentiality could retain an uncertain label.

\begin{table*}[t]
    \caption{Illustrative confidentiality categories and search anchors.
    All example facts, names, dates, and amounts are fictitious, not source excerpts
    or observed audit findings. Eligibility requires non-public, client-specific
    information supported in both texts. Categories may overlap.}
    \label{tab:audit-taxonomy}
    \centering
    \small
    \setlength{\tabcolsep}{4pt}
    \renewcommand{\arraystretch}{1.10}
    \begin{tabularx}{\textwidth}{@{}>{\raggedright\arraybackslash}p{0.22\textwidth}
        >{\raggedright\arraybackslash}X
        >{\raggedright\arraybackslash}p{0.29\textwidth}@{}}
        \toprule
        Category & Illustrative confidential fact & Candidate search anchors \\
        \midrule
        Portfolio positions and allocations &
        The client holds 1,250 units of Fund A, representing 38\% of the portfolio. &
        Product names, verified aliases, share classes, quantities, percentages \\
        Account balances and available cash &
        The client's account balance is \$420,000, including \$65,000 in available cash. &
        Amounts, account labels, balance, available cash, cash reserve \\
        Trading instructions and intentions &
        The client intends to sell \$75,000 of Bond B after 15 October while retaining Fund A. &
        Instruments, sell, redeem, retain, amounts, execution dates, conditions \\
        Transaction history &
        The client redeemed \$40,000 from Fund A on 3 September and transferred the proceeds. &
        Product, redemption, transfer, proceeds, amount, transaction date \\
        Liquidity needs and payment obligations &
        The client needs \$85,000 for a home closing in October. &
        Amount, October/Oct, home/house, closing, withdrawal, payment deadline \\
        Borrowing and pledged assets &
        The client has a \$310,000 loan secured by a position in Fund A. &
        Loan balance, collateral, pledged assets, secured lending, product names \\
        Tax-lot information and gains or losses &
        The client's Fund A holding has a cost basis of \$42 per unit and an unrealized loss of \$18,000. &
        Product, cost basis, tax lot, acquisition date, unrealized gain/loss \\
        Expected private cash flows &
        The client expects \$175,000 in business-sale proceeds in November. &
        Amount, proceeds, distribution, expected transfer, payment timing \\
        Client-specific investment constraints &
        The client requires at least \$90,000 in cash before an additional purchase. &
        Minimum cash, reserve, threshold, restriction, conditional purchase \\
        Privately negotiated financial terms &
        The client has a privately negotiated advisory fee of 0.35\%. &
        Fee rate, discount, individual agreement, quote, conditions \\
        \bottomrule
    \end{tabularx}
\end{table*}

    \textbf{Fictitious boundary example.}
    Suppose a source states that an investor holds Fund A and Bond B and plans to
    sell Bond B for an October home closing while retaining Fund A.
    A paraphrase of the sale and payment purpose warrants verification.
    Mentioning only inflation does not. Reproducing the Fund A holding can qualify
    for that fact without reproducing the full combination. Proposing to buy Bond B
    does not match the selling intention; any matching holding requires a separate
    evidence check. This example explains the rubric and does not assert that
    these details necessarily survive extraction or planning.

\subsection{Search and adjudication procedure}\label{app:audit-procedure}

    The procedure followed four stages: expert-defined taxonomy, case-specific
    anchor identification, alias-assisted manual search, and human fact verification.
    Reviewers read the synthetic text and selected informative anchors from concrete
    investor facts, such as a named instrument, amount, or dated financial event.
    Generic words such as \emph{inflation} or \emph{portfolio} were insufficient
    anchors by themselves.

    Reviewers searched across the source collection, inspected candidate passages,
    and used additional anchors to narrow candidate calls. Search guidance
    distinguished verified aliases of the same entity from broader expressions.
    Common aliases were prepared in advance; reviewers selected relevant variants
    and recorded case-specific additions as needed. Expressions such as
    \emph{sell}/\emph{liquidate}/\emph{redeem} and
    \emph{home}/\emph{house}/\emph{closing} expanded search without implying
    equivalence in every context. Different share classes, maturities, or similarly
    named products were not automatically treated as one instrument.

    Candidate verification compared relationships and material attributes in context.
    Each positive required supporting source and output spans, with a source
    identifier and location. Fragments from unrelated source calls could not be
    assembled into a purported single-source fact combination.
    Two reviewers searched and assigned initial labels independently, without
    consulting each other's judgments. Disagreements and uncertain judgments were
    discussed with a third reviewer using the collected evidence to assign final
    labels. Cases remaining ambiguous retained a final uncertain label:
    \begin{enumerate}
        \item \textbf{Confirmed match:} at least one specific non-public fact
        supported in both texts.
        \item \textbf{No match found:} the prescribed procedure was completed
        without identifying a qualifying match.
        \item \textbf{Final uncertain:} no qualifying match was confirmed, but
        semantic agreement or confidentiality remained unresolved.
    \end{enumerate}
    An output with a confirmed match counted once even when other facts remained
    uncertain or multiple matches were found.

    In the original six-strategy audit, reviewers spent approximately eight minutes
    per output, covering reading,
    anchor identification, source search, evidence verification, and recording.
    The same search protocol and resources applied to all strategy--backend settings.
    Reviewers recorded search queries, candidate source identifiers, and supporting
    spans from both texts. Completing the prescribed procedure did not imply
    exhaustive verification of every fact or every possible source match.
    This approximate duration corresponds to 160 hours per primary reviewer,
    or 320 reviewer-hours for the two independent reviews of the original 1,200 outputs;
    preparation, discussion, and third-reviewer adjudication are additional.

\subsection{Outcome, uncertainty, and final-label accounting}\label{app:audit-reporting}

    In each strategy--backend cell, $C$ denotes outputs with a confirmed match,
    $U$ those with a final uncertain label, and $N=100$ all audited outputs.
    The primary outcome is $C/N$, retaining uncertain outputs in $N$.
    Table~\ref{tab:confidentiality-audit} reports detected-match rates and Wilson
    intervals; Table~\ref{tab:audit-final-labels} reports all three final labels.
    Multiple facts and repeated ratings of an output do not increase $N$.

\begin{table}[!t]
    \caption{Detected confidential-fact matches (\%; 95\% Wilson CI).
    Each strategy--backend cell contains 100 audited outputs.
    Final uncertain cases remain in the denominator.}
    \label{tab:confidentiality-audit}
    \centering
    \footnotesize
    \setlength{\tabcolsep}{3pt}
    \renewcommand{\arraystretch}{1.08}
    \begin{tabularx}{\columnwidth}{@{}>{\raggedright\arraybackslash}Xcc@{}}
        \toprule
        Strategy & Opus~4.6 & GPT-5.4-mini \\
        \midrule
        Stochastic & 2.0 (0.6--7.0) & 2.0 (0.6--7.0) \\
        Greedy & 16.0 (10.1--24.4) & 17.0 (10.9--25.5) \\
        Top-3 & 9.0 (4.8--16.2) & 13.0 (7.8--21.0) \\
        Uniform-random & 1.0 (0.2--5.4) & 0.0 (0.0--3.7) \\
        Co-occurrence & 2.0 (0.6--7.0) & 3.0 (1.0--8.5) \\
        Source-graph conditioning & 20.0 (13.3--28.9) & 23.0 (15.8--32.2) \\
        Scenario-only prompting & 1.0 (0.2--5.4) & 1.0 (0.2--5.4) \\
        \bottomrule
    \end{tabularx}
\end{table}

\begin{table*}[t]
    \caption{Final output-level labels after adjudication; each strategy--backend
    cell contains 100 outputs. No match means no match found under the procedure,
    and uncertain denotes ambiguity remaining after adjudication.
    Totals are accounting summaries, not pooled method-performance estimates.}
    \label{tab:audit-final-labels}
    \centering
    \small
    \setlength{\tabcolsep}{5pt}
    \begin{tabular}{@{}lrrrrrr@{}}
        \toprule
        & \multicolumn{3}{c}{Opus~4.6} & \multicolumn{3}{c}{GPT-5.4-mini} \\
        \cmidrule(lr){2-4}\cmidrule(l){5-7}
        Strategy & Confirmed & No match & Uncertain & Confirmed & No match & Uncertain \\
        \midrule
        Stochastic & 2 & 98 & 0 & 2 & 96 & 2 \\
        Greedy & 16 & 84 & 0 & 17 & 80 & 3 \\
        Top-3 & 9 & 89 & 2 & 13 & 86 & 1 \\
        Uniform-random & 1 & 98 & 1 & 0 & 98 & 2 \\
        Co-occurrence & 2 & 96 & 2 & 3 & 95 & 2 \\
        Source-graph conditioning & 20 & 78 & 2 & 23 & 75 & 2 \\
        Scenario-only prompting & 1 & 98 & 1 & 1 & 99 & 0 \\
        \midrule
        Total & 51 & 641 & 8 & 59 & 629 & 12 \\
        \bottomrule
    \end{tabular}
\end{table*}

    We report two-sided 95\% Wilson score intervals without continuity correction.
    With $\widehat p=C/N$ and $z=1.96$, the interval is
    \begin{equation}
        \frac{\widehat p+z^2/(2N)
        \ \pm\ z\sqrt{\widehat p(1-\widehat p)/N+z^2/(4N^2)}}
        {1+z^2/N}.
    \end{equation}
    These marginal output-level binomial intervals describe protocol-detected matches.
    They do not account for search misses, judgment errors, or prior-estimation
    uncertainty, and do not test paired backend differences. We report descriptive
    comparisons without claims of significance or backend equivalence. Broader
    prevalence inference requires representative sampling and consideration of
    the relevant dependence structure.

    The 1,400 outputs received 110 confirmed, 1,270 no-match, and 20 uncertain labels.
    Counting all uncertain cases as positives changes stochastic's rates from
    2.0\% to 2.0\% for Opus and from 2.0\% to 4.0\% for GPT, still below the
    corresponding confirmed-only rates for greedy, top-3, and source-graph conditioning.
    For scenario-only prompting, the corresponding rates change from 1.0\% to
    2.0\% for Opus and remain 1.0\% for GPT. This limited label-sensitivity check
    is neither a confidence interval nor an upper bound on all disclosures.

    Scenario-only prompting yielded 1/100 confirmed matches per backend
    (95\% Wilson CI, 0.2--5.4\%). Its separately evaluated weighted transition JSDs
    were 0.836 and 0.812 for Opus and GPT, compared with 0.563 and 0.478 for
    stochastic sampling (Appendix~\ref{sec:eval-realism}). Both approaches had low
    observed match rates, while stochastic sampling showed closer conditional
    transition agreement. These descriptive comparisons do not establish equivalent
    confidentiality risk.

    A no-match judgment does not establish absence of confidential information:
    paraphrases, unsearched anchors, or missed passages can cause false negatives.
    Human semantic verification does not establish exhaustive search recall.
    A match does not establish that training on the matched source caused the output;
    reference search includes training and held-out calls. Uniform-random uses the
    private concept inventory and aggregate priors, so it is not source-free.
    Scenario-only prompting uses no such corpus-derived inputs and provides a
    source-free comparison under the same search protocol.

\section{Structural-overlap Assessment Details}\label{app:structural-overlap}

    This assessment screens synthetic outputs for resemblance to individual real
    records in concept content and transition structure. It identifies possible
    retention of source-specific structure for comparison across generation strategies.

\begin{table*}[t]
    \caption{Corpus-level structural-overlap assessment: 1,000 eligible recovered
    graphs per dataset--backend--strategy setting, after minimum-size filtering.
    Entity Jaccard selects the nearest real reference; transition Jaccard and LCT
    use that same reference. LCT is an unnormalized ordered subsequence length
    allowing gaps. Flagged percentage denotes entity Jaccard $\geq 0.4$.
    Bold marks the minimum within each dataset--backend block and metric.
    Lower overlap indicates less structural similarity, not a confidentiality guarantee.}
    \label{tab:structural-overlap-results}
    \centering
    \footnotesize
    \setlength{\tabcolsep}{4pt}
    \renewcommand{\arraystretch}{0.96}
    \begin{tabular*}{\textwidth}{@{\extracolsep{\fill}}lllrrrr@{}}
        \toprule
        Dataset & Backend & Strategy
        & \shortstack{Avg Entity\\Jaccard $\downarrow$}
        & \shortstack{Avg Transition\\Jaccard $\downarrow$}
        & \shortstack{Avg LCT\\$\downarrow$}
        & \shortstack{Flagged\\(\%) $\downarrow$} \\
        \midrule
        \multirow{18}{*}{\shortstack{Investor\\calls}} & \multirow{6}{*}{Opus~4.6} & Stochastic & 0.213 & 0.037 & 0.38 & 2.1 \\
         &  & Greedy & 0.314 & 0.072 & 0.62 & 20.3 \\
         &  & Top-3 & 0.296 & 0.057 & 0.51 & 13.1 \\
         &  & Uniform-random & \textbf{0.140} & \textbf{0.009} & \textbf{0.10} & \textbf{0.6} \\
         &  & Co-occurrence & 0.205 & 0.029 & 0.16 & 3.3 \\
         &  & Source-graph conditioning & 0.290 & 0.041 & 0.69 & 14.3 \\
        \cmidrule(l){2-7}
         & \multirow{6}{*}{GPT-5.4-mini} & Stochastic & 0.242 & 0.037 & 0.36 & 7.2 \\
         &  & Greedy & 0.426 & 0.207 & 0.77 & 53.2 \\
         &  & Top-3 & 0.374 & 0.126 & 0.75 & 38.4 \\
         &  & Uniform-random & \textbf{0.129} & \textbf{0.002} & \textbf{0.01} & \textbf{0.4} \\
         &  & Co-occurrence & 0.218 & 0.024 & 0.23 & 4.9 \\
         &  & Source-graph conditioning & 0.272 & 0.140 & 0.71 & 18.5 \\
        \cmidrule(l){2-7}
         & \multirow{6}{*}{Haiku~4.5} & Stochastic & 0.227 & 0.053 & 0.33 & 3.3 \\
         &  & Greedy & 0.362 & 0.129 & 0.78 & 35.9 \\
         &  & Top-3 & 0.330 & 0.076 & 0.63 & 27.2 \\
         &  & Uniform-random & \textbf{0.140} & \textbf{0.009} & \textbf{0.07} & \textbf{0.8} \\
         &  & Co-occurrence & 0.211 & 0.021 & 0.36 & 3.6 \\
         &  & Source-graph conditioning & 0.274 & 0.035 & 0.59 & 15.0 \\
        \midrule
        \multirow{18}{*}{S\&P~500} & \multirow{6}{*}{Opus~4.6} & Stochastic & 0.172 & 0.041 & 0.38 & 2.5 \\
         &  & Greedy & 0.335 & 0.059 & 0.43 & 4.5 \\
         &  & Top-3 & 0.155 & 0.036 & 0.59 & 1.1 \\
         &  & Uniform-random & \textbf{0.040} & 0.011 & \textbf{0.10} & \textbf{0.4} \\
         &  & Co-occurrence & 0.129 & \textbf{0.001} & 0.13 & 1.3 \\
         &  & Source-graph conditioning & 0.290 & 0.037 & 0.49 & 2.5 \\
        \cmidrule(l){2-7}
         & \multirow{6}{*}{GPT-5.4-mini} & Stochastic & 0.247 & 0.127 & 0.61 & 6.2 \\
         &  & Greedy & 0.426 & 0.307 & 0.87 & 43.2 \\
         &  & Top-3 & 0.404 & 0.226 & 0.75 & 32.4 \\
         &  & Uniform-random & \textbf{0.109} & \textbf{0.003} & \textbf{0.01} & \textbf{0.4} \\
         &  & Co-occurrence & 0.238 & 0.157 & 0.60 & 3.9 \\
         &  & Source-graph conditioning & 0.322 & 0.240 & 0.72 & 25.5 \\
        \cmidrule(l){2-7}
         & \multirow{6}{*}{Haiku~4.5} & Stochastic & 0.247 & 0.052 & 0.17 & 3.4 \\
         &  & Greedy & 0.382 & 0.212 & 0.42 & 29.9 \\
         &  & Top-3 & 0.330 & 0.207 & 0.63 & 25.2 \\
         &  & Uniform-random & \textbf{0.120} & \textbf{0.004} & \textbf{0.07} & \textbf{1.5} \\
         &  & Co-occurrence & 0.201 & 0.019 & 0.26 & 3.5 \\
         &  & Source-graph conditioning & 0.284 & 0.032 & 0.19 & 5.7 \\
        \bottomrule
    \end{tabular*}
\end{table*}

    \textbf{Evaluation collection and reference selection.}
    Each dataset--strategy--backend setting contains $N=1{,}000$ recovered output
    graphs after filtering out graphs with fewer than three concept nodes or two
    edges. All structural summaries use this same eligible collection. The S\&P~500
    real reference collection comprises graphs from 297 real earnings calls;
    this is distinct from the 1,000 synthetic graphs in each evaluated condition.
    Opus~4.6 is the fixed extractor for both real and synthetic texts. The reference
    pool $\mathcal{G}_{\mathrm{real}}$ includes all training and held-out real graphs.
    All strategies, including source-graph conditioning, use the same reference pool;
    reference search is not restricted to the conditioning exemplar.

    For graph $g$, let $C(g)$ denote its set of concept nodes. Define entity
    (concept-set) Jaccard and the selected reference for output $\hat g_i$ as
    \begin{align}
        J_{\mathrm{ent}}(\hat g,g)
        &= \frac{|C(\hat g)\cap C(g)|}{|C(\hat g)\cup C(g)|},\\
        g_i^* &\in \operatorname*{arg\,max}_{g\in\mathcal{G}_{\mathrm{real}}}
        J_{\mathrm{ent}}(\hat g_i,g).
    \end{align}
    The same selected reference $g_i^*$ is used for transition Jaccard and LCT.
    These are overlaps with the entity-nearest reference, not separate maxima over
    all real graphs for each metric.

    \textbf{Directed transitions and sequence order.}
    Let $\tau(g)$ be the ordered sequence of adjacent concept-transition pairs in
    the extracted concept sequence, and let $T(g)$ be its set of distinct pairs.
    Directed-transition Jaccard is
    \begin{equation}
        J_{\mathrm{trans}}(\hat g,g)
        = \frac{|T(\hat g)\cap T(g)|}{|T(\hat g)\cup T(g)|}.
    \end{equation}
    Ordered pairs preserve direction: $(u,v)$ and $(v,u)$ are distinct.
    Taking a set disregards transition sequence order and repetition. In contrast,
    \begin{equation}
        \operatorname{LCT}(\hat g,g)
        = \operatorname{LCSlen}\bigl(\tau(\hat g),\tau(g)\bigr)
    \end{equation}
    is the length of the longest common ordered subsequence of transition pairs,
    allowing gaps. It is not a contiguous substring measure. LCT is an unnormalized
    length, not a percentage, and depends on the lengths of the compared sequences.

    \textbf{Aggregation and interpretation.}
    For each of the three overlap measures $m$, we report
    $N^{-1}\sum_{i=1}^{N}m(\hat g_i,g_i^*)$. The flagged-overlap rate is
    \begin{equation}
        \frac{100}{N}\sum_{i=1}^{N}
        \mathbf{1}\!\left\{J_{\mathrm{ent}}(\hat g_i,g_i^*)\geq 0.4\right\}\%.
    \end{equation}
    The cutoff of 0.4 was initially selected by inspecting the nearest-reference
    entity-Jaccard distribution in the investor-call Opus condition and held fixed
    across datasets and backends. It is exploratory, not a calibrated confidentiality
    threshold. The small-graph filter avoids inflated scores from trivial overlap;
    it is not a confidentiality judgment and does not apply to the text-based human
    audit. Table~\ref{tab:structural-overlap-results} reports all four measures.

    Mean scores and flagged rates describe different distributional features.
    On S\&P--Opus, stochastic and source-graph conditioning both had a 2.5\%
    flagged rate despite mean entity Jaccard values of 0.172 and 0.290, respectively.
    Relative to source-graph conditioning, stochastic had lower entity Jaccard and
    LCT in all six settings and lower flagged rates in five, with this one tie;
    transition Jaccard showed a mixed comparison. These patterns describe observed
    structural overlap. Because the S\&P documents are public earnings calls,
    overlap with their company or product labels does not establish disclosure of
    non-public investor facts. Confidentiality-relevant semantic evidence is assessed
    separately by the investor-call audit in Appendix~\ref{app:confidentiality-audit}.

\section{Extended Formulation and Implementation}\label{app:extended-methods}
\subsection{Full problem formulation}

    This section defines the objects of study, namely structured financial documents
    and their graph representations, and states the privacy-motivated generation
    objective. Throughout, \emph{concept} denotes the atomic unit of financial
    meaning, represented as a vertex in the extracted graphs.

    \subsubsection{Documents, Graphs, and Realization}

    Let $\mathcal{D} = \{d_1, \ldots, d_n\}$ denote a corpus of structured financial
    documents, where each document $d_i$ is a spoken-record artifact
    such as an investor call or an earnings call. We assume a text-to-graph extraction
    operator $\mathcal{E}$ that maps each document to a directed concept graph, producing
    \begin{equation}
        G = \{g_1, \ldots, g_n\}, \quad g_i = \mathcal{E}(d_i).
    \end{equation}
    Each $g_i$ is the basic unit for prior estimation, generation, and structural
    privacy auditing. Aggregating these graphs induces a corpus-level vertex set $V$ and edge
    set $E$. The vertices in $V$ are canonical financial concepts, whose subtypes
    include topics, named entities, events, products, and company segments. Each
    directed edge $(u, v) \in E$ connects consecutive concepts in their order of
    first introduction. We further assume a graph-to-text realization
    operator $\mathcal{R}$ that approximately inverts $\mathcal{E}$, so that
    $\mathcal{R}(g_i)$ returns a document closely resembling $d_i$; it may act on a
    full graph, an ordered concept sequence, or a logically combined subset of concepts.
    Because realization is generative, $\mathcal{R}$ may also introduce grounded
    elaboration beyond the input graph, so the inversion is approximate rather than exact.

    \subsubsection{Ordered Sequences and Concept Co-occurrence}

    Two complementary views of concept structure motivate the sampling strategies
    developed later: one preserves order, and the other discards it. In the
    order-preserving view, $\mathcal{E}$ also maps each document to an ordered
    concept sequence,
    \begin{equation}
        s_i = \big(z_{i,1}, z_{i,2}, \ldots, z_{i,l_i}\big),
    \end{equation}
    where $l_i$ is the sequence length, the subscript $t$ denotes sequential
    position, $z_{i,t} \in V$, and the ordering follows the sequence in which concepts
    are first mentioned. Each $s_i$ is thus a chain-structured graph encoding
    directional discourse flow. The second view discards order through a symmetric
    co-occurrence count matrix $A$, in which $A_{jk}$ counts the documents where concepts
    $v_j$ and $v_k$ jointly appear. For sampling, we use its row-normalized form
    $\widetilde A$. The proposed method exploits this contrast between ordered
    sequence structure and the unordered structure in $A$.

    \subsubsection{Synthesis and Privacy Objective}

    The synthesis task is to generate documents that resemble those in $\mathcal{D}$
    without reproducing source-specific structure. Direct realization of the form
    $d_i'=\mathcal{R}(g_i)$ is inadvisable because a graph extracted from a single
    real document retains enough source-specific structure to raise privacy concerns.
    We therefore compare two synthesis strategies that differ in how they condition
    generation:
    \begin{enumerate}
        \item \textbf{Single-exemplar source-graph conditioning.} Sample one real graph from
        $G$ and supply it to the realizer as a structural exemplar rather than as
        source text. This strategy avoids raw source text but retains individual-graph
        conditioning.
        \item \textbf{Prior-based sampling.} Learn the priors of $\mathcal{D}$, namely
        the distributions over concepts and their orderings, and sample from these
        priors rather than from individual documents or graphs.
    \end{enumerate}
    Both strategies avoid raw source text, but only prior-based sampling removes
    individual-graph conditioning from the realization interface. The proposed
    framework instantiates the second strategy and uses source-graph conditioning
    as a comparison baseline.

\subsection{Full methodology}

    Building on this objective, DiRAFT estimates aggregate priors from document-level graphs, samples new content plans, and realizes each plan as text through a discourse layer; the generated text is re-extracted by $\mathcal{E}$ for the evaluation in Section~\ref{sec:evaluation}. The plan-producing strategies differ in how they choose successor concepts, while sharing the start-concept and plan-length priors. We present the taxonomy, prior estimation, transition policies, and discourse layer in turn. Four terms denote ordered concept structure: a \emph{sequence} $s_i$ is the ordered concepts extracted from a document; a \emph{content plan} is a freshly sampled ordered sequence of concepts; a \emph{graph plan} is that content plan instantiated as the graph $G'$ passed to the realizer; and \emph{trajectory} applies only to the diversity property. In this work, $G'$ is a node-centric realization scaffold: it represents the sampled concept sequence with directed edges between consecutive concepts.

    Algorithm~\ref{alg:diraft-synthesis} summarizes DiRAFT with stochastic sampling.
    Corpus-level priors are estimated once from training transcripts and reused
    across outputs. Here, $S$ denotes the sampled content plan, $P$ its discourse
    plan, and $\Psi$ the aggregate script priors.

\begin{algorithm}[htbp]
    \caption{DiRAFT Financial Transcript Synthesis}
    \label{alg:diraft-synthesis}
    \small
    \raggedright
    \noindent\textbf{Input:} Training transcripts $\mathcal{D}_{\mathrm{tr}}$; number of outputs $N$\par
    \noindent\textbf{Output:} Synthetic transcripts $\mathcal{D}'$\par
    \smallskip
    \renewcommand{\arraystretch}{1.08}
    \begin{tabularx}{\linewidth}{@{}r@{\quad}>{\raggedright\arraybackslash}X@{}}
    1 & Extract concept graphs and ordered concept sequences \mbox{from $\mathcal{D}_{\mathrm{tr}}$ using $\mathcal{E}$} \\
    2 & Estimate transition prior $M$, start prior $\Pi_s$, and length prior $\Pi_l$ \\
    3 & Estimate aggregate script priors $\Psi$ from $\mathcal{D}_{\mathrm{tr}}$ \\
    4 & $\mathcal{D}' \leftarrow \varnothing$ \\
    5 & \textbf{for} $j = 1,\ldots,N$: \\
    6 & \hspace*{1em}Sample $z_1 \sim \Pi_s$ and $l \sim \Pi_l$ \\
    7 & \hspace*{1em}Initialize content plan $S \leftarrow [z_1]$ \\
    8 & \hspace*{1em}\textbf{while} $|S| < l$ and $M_{\operatorname{last}(S),\cdot}$ has a positive entry: \\
    9 & \hspace*{2em}$z \leftarrow \operatorname{ConceptSampler}(\operatorname{last}(S), M)$ \\
    10 & \hspace*{2em}Append $z$ to $S$ \\
    11 & \hspace*{1em}$G' \leftarrow \operatorname{BuildGraphPlan}(S)$ \\
    12 & \hspace*{1em}$P \leftarrow \operatorname{BuildDiscoursePlan}(S, \Psi)$ \\
    13 & \hspace*{1em}$d' \leftarrow \operatorname{RealizeTranscript}(G', P)$ \\
    14 & \hspace*{1em}Add $d'$ to $\mathcal{D}'$ \\
    15 & \textbf{return} $\mathcal{D}'$ \\
    \end{tabularx}
\end{algorithm}

    \mbox{\texttt{ConceptSampler}} draws the next concept from the full row of $M$
    corresponding to the last concept in $S$. Sampling permits repeated concepts
    and stops at the target length or when no positive-probability successor exists.
    \mbox{\texttt{BuildGraphPlan}} represents the resulting $S$ as $G'$ with directed
    edges between consecutive concepts. \mbox{\texttt{BuildDiscoursePlan}} samples call goal, tone,
    speaker pattern, and turn budget from $\Psi$, distributes concepts across body
    turns, and reserves opening and closing turns. \mbox{\texttt{RealizeTranscript}}
    generates $d'$ from $G'$ and $P$, without receiving an individual source
    transcript or graph.

\subsubsection{A Taxonomy of Prior Dependence}

    Five strategies produce graph plans and vary in the transition information they
    use, from none through unordered association to the learned directed transition prior.
    Source-graph conditioning is a separate individual-conditioning baseline.

    \textbf{Uniform-random baseline (no transition prior).} Using the shared start and
    length priors defined below, it samples each remaining concept independently and
    uniformly from $V$, with replacement. Discarding transition structure, it serves
    as a low-structure reference expected to yield lower transition fidelity and higher
    trajectory diversity; it is not a formal bound on either quantity.

    \textbf{Source-graph conditioning (individual source structure as reference).} For
    each synthetic example, one graph is sampled at random from the training
    graph pool and supplied to a large language model as a single structural exemplar rather than as
    source text. The model is prompted to generate a new document in the target
    financial style, and the output is passed through $\mathcal{E}$ for evaluation. This
    strategy avoids raw source text but depends directly on the structure of individual
    real graphs and does not sample new plans from aggregate transition priors.

    \textbf{Co-occurrence sampling (unordered prior).} This strategy tests whether
    unordered concept association alone suffices. It uses the co-occurrence count matrix $A$
    to construct an undirected association graph; because $A$ is symmetric, a traversal
    positioned at $v_j$ selects the successor $v_k$ from $\widetilde A$, with probability
    proportional to their document-level co-occurrence, without asserting that $v_k$ follows $v_j$ in
    real discourse.

    \textbf{DiRAFT transition policies (ordered prior).} The three remaining strategies depend
    on the same learned directed transition prior and differ only in how they sample
    from the transition matrix $M$. \emph{The greedy sampler} chooses the most probable
    successor at each step; \emph{the top-3 sampler} restricts the choice to the three
    most probable successors and samples among them; and \emph{the stochastic sampler}
    draws each successor from the full distribution. Because these
    policies share an identical prior and pipeline, any difference in behavior is
    attributable to the sampling strategy alone.

\subsubsection{Aggregate Prior Estimation}

    We use corpus-level prior modeling to generate new content plans without
    conditioning on individual source documents or graphs. Stochastic, greedy, and top-3 sampling use the
    learned transition prior to guide concept order. Transition fidelity and structural overlap are evaluated
    empirically below.

    Three corpus-level priors are estimated: a directed transition matrix $M$, a
    start-concept distribution $\Pi_s$, and a plan-length distribution $\Pi_l$. From
    the training sequences, $\Pi_s$ records how frequently each concept appears first,
    and $\Pi_l$ records the empirical sequence-length distribution. All five
    plan-producing methods use the same $\Pi_s$ and $\Pi_l$. To estimate $M$, the
    LLM extractor returns canonical concept mentions with turn identifiers.
    Deterministic post-processing orders concepts by their first-mention turn index,
    retaining each canonical concept once to form
    $s_i = (z_{i,1}, \ldots, z_{i,l_i})$, where $z_{i,t} \in V$.
    Concepts first mentioned within the same turn are ordered by their occurrence
    in the transcript text. Directed transitions are then constructed between
    adjacent concepts in $s_i$ and aggregated across training sequences,
    \begin{equation}
        T_{uv} = \sum_i \sum_{t=1}^{l_i - 1} \mathbf{1}\{z_{i,t} = u,\ z_{i,t+1} = v\},
    \end{equation}
    and each nonzero row is normalized directly,
    \begin{equation}
        M_{uv} = \frac{T_{uv}}{\sum_{w \in V} T_{uw}}.
    \end{equation}
    Thus, unobserved transitions receive zero probability; no additive smoothing is
    used in the reported experiments. Each $M_{uv}$ is the learned probability that
    concept $v$ follows concept $u$ in the corpus-level discourse flow.

\subsubsection{Transition Policies}

    All five plan-producing methods draw $z_1 \sim \Pi_s$ and $l \sim \Pi_l$;
    previously selected concepts remain eligible, so repeats and cycles are allowed.
    For each step $t = 1, \ldots, l-1$, the transition policies sample $z_{t+1}$ from
    $M$. The top-3 sampler renormalizes over up to the three highest-probability
    successors, following stable first-appearance order when tied. The four walk-based
    methods stop at length $l$ or when no positive-probability successor is available.

    The policies expose the trade-off among privacy, realism, and diversity. We expect
    the greedy sampler to concentrate generation on the most common trajectories,
    improving local plausibility but increasing repetition and nearest-reference structural
    overlap. The stochastic sampler should instead preserve empirical discourse flow while
    introducing variation, reducing path-level copying and improving diversity, with the
    top-3 sampler occupying an intermediate position.

\begin{table*}[t]
    \caption{Discourse-layer controls and settings. Communicative controls are
    sampled from empirical priors; structural controls determine dialogue extent.}
    \label{tab:discourse-controls}
    \centering
    \small
    \setlength{\tabcolsep}{4pt}
    \renewcommand{\arraystretch}{0.88}
    \begin{tabular}{@{}>{\raggedright\arraybackslash}p{0.17\linewidth}
                    >{\raggedright\arraybackslash}p{0.35\linewidth}
                    >{\raggedright\arraybackslash}p{0.40\linewidth}@{}}
    \toprule
    Control & Description & Setting options \\
    \midrule
    \multicolumn{3}{@{}l}{\emph{Communicative framing (sampled from empirical priors)}} \\
    \textbf{Call goal} & Dialogue purpose & Initial consultation, follow-up, portfolio review, specific inquiry, relationship check \\
    \textbf{Speaker pattern} & Dialogue driver & Financial Advisor-led, Investor-led, balanced \\
    \textbf{Tone} & Communication style & Exploratory, informational, transactional, consultative \\
    \addlinespace[2pt]
    \multicolumn{3}{@{}l}{\emph{Structural extent}} \\
    \textbf{Turn budget} & Length and turn allocation & Learned from empirical data \\
    \textbf{Speaker count} & Cast size (default: two) & Configurable \\
    \bottomrule
    \end{tabular}
\end{table*}

    \subsubsection{Discourse Layer}

    The content plan specifies \emph{what} should be discussed but not
    \emph{how} the discourse should be staged. The framework therefore adds
    a lightweight discourse layer that maps the sampled content plan onto a
    document-level script, which we call the \emph{discourse plan}. This layer samples
    a call goal, tone, speaker pattern, and
    turn budget; distributes the content plan across body turns; reserves opening and
    closing turns; and enforces simple validity rules such as covering every
    mandatory concept and de-duplicating adjacent identical turns.

    Table~\ref{tab:discourse-controls} summarizes these controls, which fall into two
    groups. The first, comprising call goal, tone, and speaker pattern, governs
    communicative framing and is sampled from empirical priors. The second governs
    structural extent: the turn budget is learned from data, whereas the speaker count
    is configurable and defaults to two speakers. Because the layer conditions only on
    aggregate script priors and the sampled content plan, it provides controllable
    communicative framing without reintroducing source-fragment conditioning after the
    privacy-aware sampling step.

    \subsection{Full experimental setup}\label{app:full-setup}

    We use a proprietary corpus of 1,357 real investor-call transcripts from 846
    investors, drawn from a single month of routine business records at a financial
    institution. The transcripts discuss financial topics and products, including
    the macroeconomic outlook, investor portfolios, and investment strategies.
    Records are linked by investor identifier and month, and personally identifiable
    information (PII) is masked. The corpus is not publicly available because of
    confidentiality restrictions. Research use and access to the masked records
    were authorized by the institution, and model processing and human review
    followed its data-handling requirements. Each call is represented as a separate
    concept sequence; real-call inputs to prior estimation come only from the training
    partition. We do not model longitudinal relationships between calls.
    Training and held-out sets are disjoint at the investor level, with all calls
    from the same investor assigned to the same partition.

    We also use 297 public S\&P~500 earnings calls from February 2025, covering company
    performance, strategy, and management outlook~\citep{kurry2025sp500earnings}. The
    297 real calls provide the source documents for the S\&P graph reference
    collection. For this corpus, we use a document-level 80/20 training/held-out split.
    Both corpora's partitions were fixed before any graph aggregation and reused
    across all methods and realization backends. Except for the constructed target
    sequences in Appendix~\ref{app:single-call-effects}, all corpus-derived
    generation inputs and statistics, including the concept inventory, sampling
    priors, discourse-control priors, and source-graph exemplars, came exclusively from
    the training portion. The held-out portion served as the transition-fidelity
    reference and entered the full-collection privacy assessments only after generation.
    The real-graph UPR-2 reference was also computed from the full corpus.

    \noindent\textit{Implementation details.}
    The generation backends are Opus~4.6 and Haiku~4.5 via Amazon Bedrock and
    GPT-5.4-mini via Azure OpenAI, reported as Opus, GPT, and Haiku. Opus~4.6
    serves as the fixed extractor throughout, so that only the generation backend
    varies. Gemini~3.1 Pro performed semantic-equivalence matching and
    discourse-control identification. The coherence evaluation separately used the
    generator--judge configurations in Appendix~\ref{app:control-evaluation}.
    In the six-strategy experiments, extraction used temperature 0.0,
    whereas all three realization backends used temperature 0.3; outputs were
    capped at 8,192 tokens, and transient API
    errors were retried up to six times. Within each dataset and plan-producing
    strategy, we reuse the same graph plans and random seeds across backends. We
    evaluate the
    stochastic, greedy, and top-3 samplers, together with the uniform-random baseline,
    co-occurrence sampling, and source-graph conditioning defined in
    Section~\ref{sec:methodology}. For each of these six strategies and each
    dataset--backend setting, the structural-overlap assessment includes 1,000 recovered output graphs
    satisfying the minimum-size criteria of at least three concept nodes and two
    edges. Thus, 1,000 is the post-filter evaluation count, distinct from the
    297 real S\&P~500 source graphs.
    Scenario-only prompting is additionally evaluated for transition fidelity
    (Appendix~\ref{sec:eval-realism}) and confidential-fact detection
    (Appendix~\ref{app:confidentiality-audit}).
    The two-backend confidentiality-audit scope is specified in
    Section~\ref{sec:eval-privacy}.
    Appendix~\ref{app:prompts} provides investor-call prompt templates and explains
    the extraction-to-transition interface.

\section{Supporting Evaluation: Metrics and Full Results}\label{app:supporting-evaluation}

These evaluations assess three complementary properties: preservation of
corpus-level transition patterns, recovery of sampled plans, and repetition of
short concept paths across a synthetic collection.

\begin{table*}[t]
    \caption{Transition fidelity measured by held-out-frequency-weighted
    Jensen--Shannon divergence between synthetic and held-out real next-concept
    distributions on shared conditioning concepts. Lower values indicate closer
    conditional transition agreement; bold marks the minimum within each
    dataset--backend setting.}
    \label{tab:transition-fidelity-results}
    \centering
    \small
    \setlength{\tabcolsep}{5.25pt}
    \renewcommand{\arraystretch}{1.00}
    \begin{tabular}{@{}lcccccc@{}}
        \toprule
        & \multicolumn{3}{c}{Investor calls}
        & \multicolumn{3}{c}{S\&P~500 earnings calls} \\
        \cmidrule(lr){2-4}\cmidrule(l){5-7}
        Method & Opus & GPT & Haiku & Opus & GPT & Haiku \\
        \midrule
        \multicolumn{7}{@{}l}{\emph{Transition fidelity (weighted Jensen--Shannon divergence)}} \\
        Stochastic sampler       & \textbf{0.563} & \textbf{0.478} & \textbf{0.482} & \textbf{0.592} & \textbf{0.665} & 0.727 \\
        Greedy sampler           & 0.710 & 0.722 & 0.697 & 0.719 & 0.696 & 0.749 \\
        Top-3 sampler            & 0.649 & 0.621 & 0.611 & 0.653 & 0.674 & \textbf{0.726} \\
        Uniform-random baseline  & 0.931 & 0.988 & 0.945 & 1.000 & 1.000 & 1.000 \\
        Co-occurrence sampling   & 0.806 & 0.827 & 0.807 & 0.964 & 0.976 & 0.979 \\
        Source-graph conditioning & 0.767 & 0.771 & 0.773 & 0.861 & 0.728 & 0.941 \\
        Scenario-only prompting  & 0.836 & 0.812 & 0.754 & 0.810 & 0.826 & 0.891 \\
        \bottomrule
    \end{tabular}
\end{table*}
    \subsection{Realism: Transition Fidelity}\label{sec:eval-realism}

    Low structural overlap and few detected confidential-fact matches do not
    establish discourse fidelity. We therefore assess whether synthetic transcripts
    preserve corpus-level patterns of concept succession, using held-out real
    next-concept distributions as the reference. For a given concept, a next-concept
    distribution describes how frequently each possible successor is introduced
    next; lower divergence indicates closer agreement in these successor frequencies.

    \textit{Methods.}
    For each synthetic document, the extractor returns an ordered concept sequence,
    and aggregating adjacent transitions yields the synthetic next-concept distribution.
    Applying the same procedure to the held-out partition yields the real reference, separating
    fidelity evaluation from generation-side estimation and sampling. We report weighted
    Jensen--Shannon divergence over source concepts shared by both distributions, weighting
    each by its held-out frequency. Corpus-derived generation priors use training
    data, whereas held-out real data provide this fidelity reference. Restricting
    to shared source concepts isolates next-concept fidelity from source-concept
    coverage; lower values are better.

    We additionally evaluate scenario-only prompting to assess how closely generic
    financial task instructions reproduce the target corpus's concept-transition
    patterns. This source-free baseline specifies a financial-call scenario and leaves topic
    order to the generator, without supplying source records, extracted graphs,
    or corpus-estimated priors. This comparison evaluates the complete generation
    procedures and does not isolate the effect of the transition prior.


    \textit{Results.}
    As shown in Table~\ref{tab:transition-fidelity-results}, the stochastic
    sampler achieved the lowest weighted divergence on investor calls for all three
    backends (0.478--0.563), followed by the top-3 sampler. The S\&P~500 replication
    preserved this separation: the stochastic sampler led with Opus (0.592) and GPT
    (0.665), while the top-3 sampler narrowly led with Haiku.
    Across all backends, the transition-prior strategies achieved lower shared-source
    next-concept divergence than source-graph conditioning, co-occurrence sampling,
    and the uniform-random baseline. A broadly similar ordering was observed across
    the two financial corpora, although absolute values remained dataset-specific.

    Scenario-only prompting yielded weighted JSDs of 0.754--0.836 on investor calls
    and 0.810--0.891 on S\&P~500 earnings calls. Stochastic, greedy, and top-3
    sampling each achieved lower divergence in every corresponding dataset--backend
    setting. Stochastic sampling's scores were lower by 0.161--0.334.

    Scenario-only prompting nevertheless achieved lower divergence than
    uniform-random sampling in all six settings and co-occurrence sampling in five.
    Across both corpora and all three backends, the transition-prior strategies
    achieved closer agreement with held-out next-concept distributions than the
    other evaluated strategies.

    \subsection{Recoverability: Content Preservation}\label{sec:eval-recoverability}

    A graph plan is useful only if its intended content survives realization, so we
    measure how much planned content remains recoverable after a document is realized
    and re-extracted, while allowing document-grounded enrichment.

    \textit{Methods.}
    For each example, a
    graph plan and its discourse plan are realized into a document and re-extracted
    into a recovered graph. Nodes are aligned by exact or substring matching, by
    Jaccard overlap (intersection over union) of lemmatized token sets at a $0.7$
    threshold, and, for unresolved pairs, by Gemini~3.1 Pro semantic-equivalence matching.
    After node alignment, edge recoverability compares untyped directed pairs of
    adjacent concepts in the sampled plan and the recovered first-introduction sequence. Source-graph conditioning is excluded because it produces no
    paired plan. For each document, recall is the fraction of planned items with at
    least one aligned recovered counterpart, whereas grounding-adjusted precision is
    the fraction of recovered items that either align with the plan or are supported
    in the generated document. Node and edge F1 are macro-averaged across documents.

    \textit{Results.}
    As Table~\ref{tab:recoverability-results} shows, most planned content was preserved
    with limited round-trip loss: across both datasets and all backends, most node and
    edge F1 scores were at least 0.90, and the lowest observed value was 0.886. The
    greedy sampler and uniform-random baseline also achieved high recoverability,
    confirming that this property evaluates the execution of a given plan rather than
    overall generator quality.

\begin{table*}[t]
    \caption{Grounding-adjusted node and edge recoverability (macro-F1) across
    datasets and backends; higher is better.}
    \label{tab:recoverability-results}
    \centering
    \small
    \setlength{\tabcolsep}{5.25pt}
    \renewcommand{\arraystretch}{1.00}
    \begin{tabular}{@{}lcccccc@{}}
        \toprule
        & \multicolumn{3}{c}{Investor calls}
        & \multicolumn{3}{c}{S\&P~500 earnings calls} \\
        \cmidrule(lr){2-4}\cmidrule(l){5-7}
        Method & Opus & GPT & Haiku & Opus & GPT & Haiku \\
        \midrule
        \multicolumn{7}{@{}l}{\emph{Node recoverability (macro-averaged node F1)}} \\
        Stochastic sampler      & 0.912 & 0.986 & 0.958 & 0.886 & 0.974 & 0.940 \\
        Greedy sampler          & 0.909 & 0.992 & 0.964 & 0.963 & 0.995 & 0.981 \\
        Top-3 sampler           & 0.914 & 0.986 & 0.962 & 0.900 & 0.984 & 0.936 \\
        Uniform-random baseline & 0.918 & 0.990 & 0.963 & 0.964 & 0.994 & 0.982 \\
        Co-occurrence sampling  & 0.915 & 0.987 & 0.963 & 0.893 & 0.987 & 0.944 \\
        \multicolumn{7}{@{}l}{\emph{Edge recoverability (macro-averaged edge F1)}} \\
        Stochastic sampler      & 0.918 & 0.987 & 0.962 & 0.887 & 0.974 & 0.940 \\
        Greedy sampler          & 0.918 & 0.993 & 0.968 & 0.964 & 0.995 & 0.981 \\
        Top-3 sampler           & 0.921 & 0.987 & 0.965 & 0.901 & 0.985 & 0.937 \\
        Uniform-random baseline & 0.923 & 0.991 & 0.965 & 0.964 & 0.994 & 0.982 \\
        Co-occurrence sampling  & 0.920 & 0.988 & 0.965 & 0.893 & 0.986 & 0.944 \\
        \bottomrule
    \end{tabular}
\end{table*}
    \subsection{Diversity: Trajectory Anti-Collapse}\label{sec:eval-diversity}

    Transition fidelity alone does not rule out repeated use of a few short concept
    paths. We therefore assess trajectory diversity as a safeguard against collapse
    rather than an objective to maximize.

    \textit{Methods.}
    We report the two-step unique path ratio
    (UPR-2). Let $\mathcal{P}_2$ be the multiset of all directed length-two subpaths
    pooled across an evaluated graph collection. We define
    \begin{equation}
        \operatorname{UPR\text{-}2}
        = \frac{\left|\operatorname{uniq}(\mathcal{P}_2)\right|}
        {\left|\mathcal{P}_2\right|}.
    \end{equation}
    The numerator counts distinct directed subpaths, whereas the denominator counts
    all occurrences; reversing a subpath therefore gives a different element. Graphs
    with fewer than three concepts contribute to neither count. Although co-occurrence
    sampling uses an undirected prior during generation, its outputs are evaluated
    using the same directed criterion. Lower values indicate greater collapse. Within
    each dataset, graph plans are shared across realization backends; we therefore
    compute plan-level UPR-2 once
    per method on the complete plan bank and recovered-graph UPR-2 separately for each
    backend. We interpret each dataset against its own full-corpus real-graph UPR-2
    reference (approximately 0.99).
    As before, source-graph conditioning is excluded.

\begin{table}[!t]
    \caption{UPR-2 on shared investor-call plans and backend-specific
    recoveries; higher means less collapse (real graphs: $\approx 0.99$).}
    \label{tab:diversity-results}
    \centering
    \small
    \setlength{\tabcolsep}{1.5pt}
    \begin{tabular}{@{}lcccc@{}}
        \toprule
        & \multicolumn{1}{c}{Planned $G'$}
        & \multicolumn{3}{c}{Recovered $\widehat G$} \\
        \cmidrule(lr){2-2}\cmidrule(lr){3-5}
        Method & Shared & Opus & GPT & Haiku \\
        \midrule
        Stochastic sampler      & 0.937 & 0.971 & 0.964 & 0.969 \\
        Greedy sampler          & 0.284 & 0.722 & 0.417 & 0.654 \\
        Top-3 sampler           & 0.467 & 0.807 & 0.623 & 0.756 \\
        Uniform-random baseline & 1.000 & 0.995 & 1.000 & 0.990 \\
        Co-occurrence sampling  & 0.998 & 0.990 & 0.996 & 0.990 \\
        \bottomrule
    \end{tabular}
\end{table}

    \textit{Results.}
    Table~\ref{tab:diversity-results} reports the investor-call results; the same
    qualitative pattern held for S\&P~500, whose values are omitted for space. The
    greedy sampler consistently produced the lowest
    UPR-2 and remained well below the reference even after realization, indicating
    clear trajectory collapse. The top-3 sampler showed a milder concentration
    pattern, while the stochastic sampler, uniform-random baseline, and co-occurrence
    sampling remained close to the reference.

\section{Discourse-control Evaluation}\label{app:control-evaluation}
    We test whether changing discourse settings produces distinguishable outputs
    and whether adding the discourse layer affects coherence.
    The discourse layer is configurable: call goal, tone, and speaker pattern default
    to empirical priors but can be overridden.

    \textit{Methods.}
    On investor-call data, Opus generated
    50 fixed-plan counterfactual pairs per control, each differing only in setting. A
    blinded Gemini~3.1 Pro judge assigned the two outputs to two candidate settings in a two-way
    forced-choice task (50\% chance). For
    coherence, blinded judges rated 100 matched with/without-layer pairs per
    generator$\rightarrow$judge configuration (Opus$\rightarrow$Opus,
    Opus$\rightarrow$GPT, GPT$\rightarrow$Opus) on a 1--5 scale (higher is better).

    \textit{Results.}
    Forced-choice setting identification yielded 97.3\% macro-averaged accuracy.
    Adding the layer reduced mean coherence by 0.12--0.73 points, indicating a
    model-judged controllability--coherence trade-off in these settings.

\section{Human Evaluation of Concept and Transition Extraction}\label{app:extraction-validation}

\textbf{Objective and sample.}
Because graph-based comparisons depend on the extractor, we assess how accurately
it recovers concepts and their first-mention transitions against human reference
annotations.
We evaluated the fixed Opus~4.6 extractor on 600 investor-call transcripts:
60 held-out real calls and 30 synthetic transcripts per strategy--backend
combination. Synthetic texts covered stochastic, greedy, top-3, uniform-random,
co-occurrence, and source-graph conditioning, using Opus~4.6, GPT-5.4-mini,
and Haiku~4.5 as generation backends. The same annotation criteria applied to all texts.

\textbf{Annotation criteria and procedure.}
Two annotators constructed reference annotations from the transcript text.
Annotation took approximately 45 hours per annotator (90 person-hours in total).
The reference representation comprised canonical financial concepts and their
order of first mention. Concepts included topics, named entities, products,
and events under the extraction schema. Synonymous expressions denoting the
same concept received a common canonical label, while related but distinct
concepts remained separate. Each concept annotation included a supporting text
span and its first-mention location. Repeated mentions did not introduce
additional concepts. Directed transitions connected adjacent concepts in the
first-mention sequence; reversing a pair changed the transition. The finalized
reference annotations were compared with the fixed extractor output.

\textbf{Matching and metrics.}
Predicted concepts were aligned one-to-one with reference concepts under the
canonicalization rules. Concept precision measured the fraction of predicted
concepts matched to the reference, and concept recall measured the fraction
of reference concepts recovered. Transition precision and recall were computed
over sets of adjacent directed concept pairs using the same concept alignment.
A transition matched only when both endpoints and their direction agreed.
Transitions involving unmatched concepts remained in the respective denominators.
We compute concept and directed-transition precision and recall separately for
each transcript, then report unweighted macro averages within each evaluation
group. Each transcript contributes equally to the reported scores.

\textbf{Results.}
Table~\ref{tab:extraction-validation} reports all 19 evaluation groups.
On real investor calls, concept precision and recall were 94.6\% and 92.3\%,
respectively, while transition precision and recall were 91.3\% and 88.1\%.
Across synthetic conditions, concept precision ranged from 92.4--95.6\% and
recall from 91.4--95.5\%; transition precision ranged from 88.2--91.9\% and
recall from 87.6--91.7\%. These results provide empirical support for the
overall accuracy of the extractor on the evaluated investor-call texts.

\begin{table*}[t]
    \caption{Concept and transition extraction against human reference annotations
    on investor-call texts. Backend denotes the generation model; Opus~4.6 is the
    fixed extractor throughout. The sample contains 60 real transcripts and 30
    synthetic transcripts per strategy--backend combination. Values are
    transcript-level macro-averaged precision and recall percentages.}
    \label{tab:extraction-validation}
    \centering
    \small
    \setlength{\tabcolsep}{4pt}
    \renewcommand{\arraystretch}{1.05}
    \begin{tabular*}{\textwidth}{@{\extracolsep{\fill}}llrrrr@{}}
        \toprule
        Method / Text source & Backend
        & \multicolumn{2}{c}{Concept} & \multicolumn{2}{c}{Transition} \\
        \cmidrule(lr){3-4}\cmidrule(l){5-6}
        & & Precision $\uparrow$ & Recall $\uparrow$
        & Precision $\uparrow$ & Recall $\uparrow$ \\
        \midrule
        Original real & --- & 94.6 & 92.3 & 91.3 & 88.1 \\
        \midrule
        Stochastic & Opus & 95.0 & 95.5 & 91.8 & 91.6 \\
        Greedy & Opus & 95.6 & 94.9 & 91.9 & 91.7 \\
        Top-3 & Opus & 94.2 & 92.2 & 91.0 & 88.6 \\
        Uniform-random & Opus & 92.4 & 91.8 & 88.2 & 87.9 \\
        Co-occurrence & Opus & 93.5 & 93.2 & 90.0 & 89.4 \\
        Source-graph conditioning & Opus & 94.7 & 92.6 & 89.7 & 88.7 \\
        \midrule
        Stochastic & GPT & 94.3 & 92.5 & 91.0 & 88.2 \\
        Greedy & GPT & 95.2 & 93.5 & 91.2 & 90.3 \\
        Top-3 & GPT & 93.7 & 92.7 & 90.0 & 88.9 \\
        Uniform-random & GPT & 93.6 & 93.4 & 89.0 & 90.1 \\
        Co-occurrence & GPT & 93.7 & 93.3 & 90.2 & 88.4 \\
        Source-graph conditioning & GPT & 95.2 & 92.4 & 90.2 & 88.6 \\
        \midrule
        Stochastic & Haiku & 93.7 & 91.7 & 89.5 & 88.0 \\
        Greedy & Haiku & 94.6 & 93.9 & 90.7 & 88.8 \\
        Top-3 & Haiku & 93.9 & 91.9 & 89.6 & 88.3 \\
        Uniform-random & Haiku & 93.6 & 91.4 & 88.6 & 87.6 \\
        Co-occurrence & Haiku & 94.2 & 93.0 & 90.4 & 88.7 \\
        Source-graph conditioning & Haiku & 94.2 & 92.2 & 89.8 & 89.0 \\
        \bottomrule
    \end{tabular*}
\end{table*}

\section{Human Preference for Topic Progression}\label{app:order-evaluation}

This comparison examines whether preserving the sampled order of concepts matters
to human preference for topic progression.

\subsection{Comparison and categorical votes}
Two annotators independently evaluated 100 pairs of investor-call transcripts for
 each backend, Opus~4.6 and GPT-5.4-mini. Each pair compared an output generated
from an original stochastic plan with an output generated from its order-shuffled
counterpart. Annotators provided categorical preference votes for topic progression,
without assigning numerical ratings. We report a vote as \emph{Original} when the
transcript generated from the original stochastic plan was preferred, and as
\emph{Shuffled} when its counterpart was preferred. \emph{Tie} indicates no
preference between the two transcripts. Original denotes a generation condition,
not a real source transcript.


\subsection{Vote aggregation and uncertainty}
We report the counts and proportions of all three vote categories. As a summary,
we calculate tie-adjusted preference for Original as
\begin{equation}
    \widehat p = \frac{N_{\mathrm{Original}}+\tfrac12 N_{\mathrm{Tie}}}
    {N_{\mathrm{votes}}},\qquad N_{\mathrm{votes}}=200,
\end{equation}
where each backend contributes two votes for each of 100 pairs. Giving Tie votes
half weight is an analysis convention, not an annotator-assigned numerical rating;
50\% denotes no aggregate preference. Pointwise 95\% percentile-bootstrap
confidence intervals resample intact pairs, retaining both annotators' votes and
recalculating the summary. We evaluate the empirical bootstrap distribution by
convolution without Monte Carlo sampling. The intervals describe pair-sampling
uncertainty for the observed two-annotator panel. We also report each annotator's
vote distribution, observed agreement, and unweighted Cohen's $\kappa$ calculated
directly from the categorical votes.

\begin{table*}[t]
    \caption{Human preference for topic progression on investor calls, with
    100 transcript pairs and 200 categorical votes per backend. Original and
    Shuffled denote the preferred generation condition; Tie denotes no preference.
    Tie-adjusted preference is the proportion of Original votes plus half the
    proportion of Tie votes. The 95\% percentile-bootstrap CIs retain both votes
    together when resampling pairs.}
    \label{tab:order-preference}
    \centering
    \small
    \setlength{\tabcolsep}{5pt}
    \begin{tabular}{@{}lcccc@{}}
        \toprule
        Backend & Original preferred & Tie & Shuffled preferred &
        \shortstack{Tie-adjusted preference\\(95\% CI)} \\
        \midrule
        Opus~4.6 & 97/200 (48.5\%) & 65/200 (32.5\%) & 38/200 (19.0\%)
        & 64.75\% [58.00\%, 71.25\%] \\
        GPT-5.4-mini & 90/200 (45.0\%) & 67/200 (33.5\%) & 43/200 (21.5\%)
        & 61.75\% [55.25\%, 68.00\%] \\
        \bottomrule
    \end{tabular}
\end{table*}

\subsection{Preferences and inter-annotator agreement}
Each annotator cast more Original than Shuffled votes for both backends
(Table~\ref{tab:order-agreement}). Tie-adjusted preferences derived from their votes
were 65.0\% and 64.5\% for Opus~4.6, and 64.5\% and 59.0\% for GPT-5.4-mini.
Ties accounted for 32.5\% and 33.5\% of votes, respectively
(Table~\ref{tab:order-preference}). Inter-annotator agreement was 71\% for Opus~4.6
and 61\% for GPT-5.4-mini, with unweighted Cohen's $\kappa$ values of 0.537 and 0.391.
These observations support a preference for outputs from the original stochastic
plans in the evaluated comparison. They do not establish superiority over other
ordering methods or over co-occurrence sampling in human evaluation.

\begin{table*}[t]
    \caption{Individual votes and inter-annotator agreement for topic progression.
    Each annotator cast one categorical vote for each of 100 pairs per backend.
    Observed agreement is the proportion of pairs receiving identical labels;
    expected chance agreement uses the annotators' marginal label frequencies.
    Cohen's $\kappa$ is unweighted. Direct disagreements indicate one Original
    vote and one Shuffled vote for the same pair.}
    \label{tab:order-agreement}
    \centering
    \small
    \setlength{\tabcolsep}{6pt}
    \begin{tabular}{@{}lcc@{}}
        \toprule
        Metric & Opus~4.6 & GPT-5.4-mini \\
        \midrule
        Annotator 1: Original / Tie / Shuffled & 51 / 28 / 21 & 47 / 35 / 18 \\
        Annotator 2: Original / Tie / Shuffled & 46 / 37 / 17 & 43 / 32 / 25 \\
        Annotator 1: Tie-adjusted preference & 65.0\% & 64.5\% \\
        Annotator 2: Tie-adjusted preference & 64.5\% & 59.0\% \\
        Observed agreement & 71/100 (71.0\%) & 61/100 (61.0\%) \\
        Expected chance agreement & 37.39\% & 35.91\% \\
        Unweighted Cohen's $\kappa$ & 0.537 & 0.391 \\
        Direct Original--Shuffled disagreements & 6/100 (6.0\%) & 12/100 (12.0\%) \\
        \bottomrule
    \end{tabular}
\end{table*}

\section{Generated Transcript Preference}\label{app:strategy-preference-evaluation}

Aggregate transition similarity does not establish whether a complete dialogue
has clear topic connections, motivated shifts, or a plausible financial discussion.
This evaluation complements transition-fidelity statistics with human preferences
between generated investor-call transcripts.

\subsection{Comparisons and annotation records}
We compared stochastic sampling with
scenario-only prompting, source-graph conditioning, and co-occurrence sampling,
using Opus~4.6 for generation. Source-graph conditioning supplies an individual
training graph, while co-occurrence sampling uses aggregate unordered concept
co-occurrence counts (Section~\ref{sec:methodology}). Scenario-only prompting was
included as an additional comparison for text-level quality. It is also evaluated
for transition fidelity in Appendix~\ref{sec:eval-realism}.

Each comparison contains 100 evaluated transcript pairs. Two annotators each
judged the same set of 100 pairs, yielding 200 categorical judgments per
comparison. Across the three comparisons, each annotator supplied 300 judgments,
for 600 judgments over 300 pair comparisons. These counts describe comparisons
and judgments, not necessarily distinct transcripts. The recorded categories are
\emph{Stochastic preferred}, \emph{Tie} (no clear preference), and
\emph{Baseline preferred}. Table~\ref{tab:strategy-preference-joint} gives the
joint counts, with Annotator~1 defining rows and Annotator~2 defining columns.


\subsection{Preference metric and disagreement handling}
We retain both recorded judgments for every pair, including disagreements,
without constructing a consensus label. A disagreement between a Stochastic
vote and a Baseline vote therefore remains two opposing votes, not two Tie votes.
We report all three category counts and calculate tie-adjusted preference as
\begin{equation}
    \widehat p =
    \frac{N_{\mathrm{Stochastic}}+\tfrac12 N_{\mathrm{Tie}}}
    {N_{\mathrm{judgments}}},\qquad N_{\mathrm{judgments}}=2\times100=200.
\end{equation}
Giving ties half weight is an analysis convention rather than an
annotator-assigned numerical rating. A value of 50\% indicates no aggregate
preference. The estimate is also the mean of the 100 pair-average scores,
where each categorical judgment receives a score of 1, 0.5, or 0. The estimate
is not the proportion of unanimously won pairs or a percentage increase in quality.

Percentile-bootstrap intervals resample intact pairs, retaining both judgments
together. We evaluate the empirical bootstrap distribution by convolution without
Monte Carlo sampling. The three comparisons form the multiplicity family:
Bonferroni-adjusted intervals use the $0.05/(2\times3)$ and
$1-0.05/(2\times3)$ quantiles, corresponding to individual nominal coverage of
98.33\% and approximately 95\% simultaneous coverage. Bootstrap coverage remains
approximate. This analysis assumes independently sampled pairs within each
comparison and conditions on the observed two-annotator panel; it does not treat
the 200 judgments as independent samples or estimate variation across a
population of annotators. Repeated outputs or other shared generation units
within a comparison would require accounting for that dependence.

We assess agreement using the proportion of pairs receiving identical labels and
unweighted Cohen's $\kappa$, with expected agreement calculated from the two
annotators' marginal category frequencies. We also report each annotator's
preference and the number of pairs with directly opposing votes.

\subsection{Preference results}
Stochastic outputs received tie-adjusted preferences of 75.25\%, 72.50\%, and
62.75\% against scenario-only prompting, source-graph conditioning, and co-occurrence,
respectively (Table~\ref{tab:strategy-preference-results}). All three adjusted
intervals have lower bounds above 50\%. The corresponding pointwise 95\%
intervals are 69.75--80.50\%, 66.75--78.00\%, and 56.00--69.50\%.
Both annotators individually favored stochastic outputs in every comparison.

\begin{table*}[t]
    \caption{Human preference across generation strategies with Opus~4.6.
    Each comparison contains 100 transcript pairs, each judged by both annotators:
    100 judgments per annotator and $N_{\mathrm{judgments}}=200$ in total.
    Counts and percentages describe judgments. Tie-adjusted preference gives ties
    half weight. Intervals retain both judgments when resampling pairs and use
    Bonferroni adjustment across the three comparisons, targeting approximately
    95\% simultaneous coverage.}
    \label{tab:strategy-preference-results}
    \centering
    \small
    \setlength{\tabcolsep}{4pt}
    \begin{tabular}{@{}lcccc@{}}
        \toprule
        Baseline & Stochastic preferred & Tie & Baseline preferred &
        \shortstack{Tie-adjusted preference\\(adjusted interval)} \\
        \midrule
        Scenario-only prompting & 119/200 (59.5\%) & 63/200 (31.5\%) & 18/200 (9.0\%)
        & 75.25\% [68.50\%, 81.50\%] \\
        Source-graph conditioning & 112/200 (56.0\%) & 66/200 (33.0\%) & 22/200 (11.0\%)
        & 72.50\% [65.50\%, 79.00\%] \\
        Co-occurrence & 97/200 (48.5\%) & 57/200 (28.5\%) & 46/200 (23.0\%)
        & 62.75\% [54.50\%, 70.75\%] \\
        \bottomrule
    \end{tabular}
\end{table*}

\subsection{Agreement and interpretation}
Exact agreement was 57\%, 55\%, and 59\%, with corresponding $\kappa$ values
of 0.202, 0.204, and 0.352 (Table~\ref{tab:strategy-preference-agreement}).
Directly opposing votes occurred in 4, 5, and 10 pairs, respectively. The other
39, 40, and 31 disagreements involved one Tie and one directional preference.
Thus, the aggregate preference favored stochastic sampling despite disagreement
on individual pairs. The results do not establish high inter-annotator agreement.

These findings support a preference for stochastic outputs under the evaluated
annotation task, for Opus~4.6 and this two-annotator panel. This comparison does
not isolate the effect of concept order; the order ablation is reported separately
in Appendix~\ref{app:order-evaluation}. Nor does it establish that adding the
discourse layer improves coherence, or that lower JSD causes higher human
preference. Financial internal consistency, externally verified factual accuracy,
conversational naturalness, and confidentiality protection require their own
evidence and are not separate outcomes of these preference counts.

\begin{table*}[t]
    \caption{Individual judgments and agreement for the strategy comparisons.
    S/T/B denote Stochastic preferred, Tie, and Baseline preferred. Each annotator
    judged 100 pairs per comparison. Agreement and disagreement counts use pairs
    as the denominator; Cohen's $\kappa$ is unweighted.}
    \label{tab:strategy-preference-agreement}
    \centering
    \small
    \setlength{\tabcolsep}{6pt}
    \begin{tabular}{@{}lccc@{}}
        \toprule
        Metric & \shortstack{Scenario-only\\prompting} & \shortstack{Source-graph\\conditioning} & Co-occurrence \\
        \midrule
        Annotator 1: S / T / B & 59 / 33 / 8 & 57 / 32 / 11 & 52 / 27 / 21 \\
        Annotator 2: S / T / B & 60 / 30 / 10 & 55 / 34 / 11 & 45 / 30 / 25 \\
        Annotator 1: Tie-adjusted preference & 75.5\% & 73.0\% & 65.5\% \\
        Annotator 2: Tie-adjusted preference & 75.0\% & 72.0\% & 60.0\% \\
        Exact agreement & 57/100 (57\%) & 55/100 (55\%) & 59/100 (59\%) \\
        Expected chance agreement & 46.10\% & 43.44\% & 36.75\% \\
        Unweighted Cohen's $\kappa$ & 0.202 & 0.204 & 0.352 \\
        Directly opposing votes & 4/100 & 5/100 & 10/100 \\
        Tie versus directional preference & 39/100 & 40/100 & 31/100 \\
        \bottomrule
    \end{tabular}
\end{table*}

\begin{table}[t]
    \caption{Joint categorical counts for all three strategy comparisons.
    Each block is a $3\times3$ table: rows give Annotator~1's choice and columns
    give Annotator~2's choice. Baseline refers to the method named in that block.
    Every pair contributes to exactly one cell, and each block sums to 100 pairs.}
    \label{tab:strategy-preference-joint}
    \centering
    \fontsize{7.5}{9.2}\selectfont
    \setlength{\tabcolsep}{3pt}
    \begin{tabular}{@{}llrrr@{}}
        \toprule
        & & \multicolumn{3}{c}{Annotator 2} \\
        \cmidrule(l){3-5}
        Baseline & Annotator 1 & Stochastic & Tie & Baseline \\
        \midrule
        \multirow{3}{*}{\shortstack[l]{Scenario-only\\prompting}} & Stochastic & 41 & 16 & 2 \\
        & Tie & 17 & 12 & 4 \\
        & Baseline & 2 & 2 & 4 \\
        \midrule
        \multirow{3}{*}{Source-graph conditioning} & Stochastic & 37 & 17 & 3 \\
        & Tie & 16 & 13 & 3 \\
        & Baseline & 2 & 4 & 5 \\
        \midrule
        \multirow{3}{*}{Co-occurrence} & Stochastic & 33 & 12 & 7 \\
        & Tie & 9 & 13 & 5 \\
        & Baseline & 3 & 5 & 13 \\
        \bottomrule
    \end{tabular}
\end{table}

\section{Single-Call Effects}\label{app:single-call-effects}

\begin{table}[t]
    \caption{Target reproduction in plans, averaged across 30 targets
    with 200 outputs per target, method, and condition.
    Before and After are percentages; approximate increases in percentage points
    are calculated from the displayed means.}
    \label{tab:single-call-plans}
    \centering
    \fontsize{7.5}{9.2}\selectfont
    \setlength{\tabcolsep}{2pt}
    \renewcommand{\arraystretch}{1.08}
    \begin{tabular*}{\columnwidth}{@{\extracolsep{\fill}}llrrr@{}}
        \toprule
        Method & Measure & Before & After & \shortstack{Approximate\\increase (pp)} \\
        \midrule
        DiRAFT & Full-path reproduction & 0.00 & 0.20 & +0.20 \\
        Source-path conditioning & Full-path reproduction & 0.00 & 1.25 & +1.25 \\
        \addlinespace
        DiRAFT & Transition coverage & 0.00 & 0.25 & +0.25 \\
        Source-path conditioning & Transition coverage & 0.00 & 1.25 & +1.25 \\
        \addlinespace
        DiRAFT & Concept coverage & 8.50 & 9.34 & +0.84 \\
        Source-path conditioning & Concept coverage & 8.50 & 9.54 & +1.04 \\
        \bottomrule
    \end{tabular*}
\end{table}

\begin{table}[t]
    \caption{Target reproduction after re-extraction, averaged across 30 targets
    with 200 outputs per target, method, and condition.
    Before and After are percentages; approximate increases in percentage points
    are calculated from the displayed means.}
    \label{tab:single-call-reextraction}
    \centering
    \fontsize{7.5}{9.2}\selectfont
    \setlength{\tabcolsep}{2pt}
    \renewcommand{\arraystretch}{1.08}
    \begin{tabular*}{\columnwidth}{@{\extracolsep{\fill}}llrrr@{}}
        \toprule
        Method & Measure & Before & After & \shortstack{Approximate\\increase (pp)} \\
        \midrule
        DiRAFT & Full-path reproduction & 0.02 & 0.12 & +0.10 \\
        Source-path conditioning & Full-path reproduction & 0.02 & 0.83 & +0.81 \\
        \addlinespace
        DiRAFT & Transition coverage & 0.35 & 0.43 & +0.08 \\
        Source-path conditioning & Transition coverage & 0.35 & 1.08 & +0.73 \\
        \addlinespace
        DiRAFT & Concept coverage & 7.25 & 7.55 & +0.30 \\
        Source-path conditioning & Concept coverage & 7.25 & 8.28 & +1.03 \\
        \bottomrule
    \end{tabular*}
\end{table}

\textbf{Objective.}
Structural Overlap measures resemblance to individual source records, while
Confidential-Fact Detection checks whether generated transcripts reproduce specific
non-public investor facts. Neither assessment isolates the change caused by adding
one call representation. This controlled in/out experiment measures that change by
adding a constructed concept sequence and comparing its reproduction before and
after inclusion.

\textbf{Methods.}
We used concept sequences extracted from 100 real investor calls in the training set as the background pool.
Each sequence followed the order in which concepts first appeared in its source
call. We constructed 30 target sequences using concepts already present in this
pool, with every adjacent target transition absent from the background.

Each target was tested separately. The Before condition used the 100 background
sequences; the After condition added one occurrence of the target sequence.
DiRAFT re-estimated its transition, starting-concept, and sequence-length priors
and used stochastic sampling to generate new plans. Source-path conditioning
uniformly selected one complete sequence from the corresponding pool. This
baseline is a sequence-based version of the source-graph conditioning approach
in Section~\ref{sec:methodology}.

Opus~4.6 performed both transcript generation and re-extraction. All other
generation settings remained unchanged across methods and between Before and
After. We evaluated 200 outputs per target, method, and condition, assessing
both the conditioning plan and the re-extracted concept sequence for each output.
Across 30 targets, each method--condition summary comprises 6,000 target-specific
output evaluations.


\textbf{Reproduction measures.}
We evaluated reproduction at three levels. \emph{Full-path reproduction} is the
percentage of outputs containing the complete target sequence in the correct
order, with its concepts adjacent. \emph{Transition coverage} is the average
fraction of target transitions present in each output. \emph{Concept coverage}
is the average fraction of target concepts present in each output. Repeated
occurrences of the same target item within an output did not increase its score.

Plan matching used the sampled or selected sequence order. Re-extracted paths
used the first-appearance order of canonical concepts in the generated transcript.
Transition matches required the correct endpoints and direction. Intervening
concepts were retained when checking adjacency, so merely mentioning all target
concepts did not qualify as full-path reproduction.

\textbf{Human verification.}
Two annotators checked the target concept matches and their first-appearance
order against the generated transcripts, including missed concepts, unsupported
matches, and ordering errors. We followed the canonicalization and ordering
criteria described in Appendix~\ref{app:extraction-validation}. That evaluation
separately reports the extractor's concept and transition accuracy.

\textbf{Aggregation and results.}
Each output was evaluated against the target assigned to its experiment. We
averaged scores within each target and calculated its change from Before to After.
The Before, After, and change summaries give equal weight to each of the 30 targets.

Tables~\ref{tab:single-call-plans} and~\ref{tab:single-call-reextraction} report the
plan and re-extraction results. DiRAFT showed smaller average increases in
full-path reproduction, transition coverage, and concept coverage at both stages.
These findings support lower sensitivity to the inclusion of an individual
concept sequence in the evaluated setting. Because the intervention adds a
sequence rather than a raw transcript, it measures structural influence rather
than confidential-fact disclosure.


\section{Investor-Call Prompt Templates}\label{app:prompts}

The templates below document the investor-call extraction, realization, baseline,
and model-judging tasks. They are normalized for presentation, with standardized
participant labels and clarified instructions and formatting, rather than
verbatim execution logs. Investor denotes the client receiving financial guidance;
Financial Advisor denotes the other participant. Braced placeholders indicate
task inputs. Similar prompts with minor adaptations were used for the S\&P~500
experiments; those variants are omitted for conciseness.

\newenvironment{promptblock}{%
    \setlength{\fboxsep}{5pt}\setlength{\fboxrule}{0.35pt}%
    \setlength{\OuterFrameSep}{5pt}%
    \def\FrameCommand{\fcolorbox{black!30}{black!1}}%
    \MakeFramed{\advance\hsize-\width\FrameRestore}%
    \fontsize{7.5}{9.2}\selectfont\raggedright
    \setlength{\parindent}{0pt}\setlength{\parskip}{2pt}%
    \setlength{\topsep}{3pt}\setlength{\partopsep}{0pt}%
    \clubpenalty=10000\widowpenalty=10000
}{\par\endMakeFramed}

\subsection{Relationship extraction}\label{app:prompt-extraction}

The fixed Opus~4.6 extractor returns evidence-linked mention records.
Post-processing uses the returned canonical concepts to construct their
first-introduction sequence and adjacent transitions as described in
Appendix~\ref{app:extended-methods}. Speaker-to-concept \texttt{DISCUSSED} records
identify mention provenance and are distinct from these concept transitions.
Verbatim evidence is retained for traceability and human evaluation, including
extractor validation, and is excluded from the inputs supplied to the realizer.
The internal origin label \texttt{crew\_introduced} is displayed as
\texttt{advisor\_introduced} below; verbatim transcript evidence retains its
original wording and speaker labels.

\begin{promptblock}
\noindent\textbf{System prompt.}\par\nobreak
You are a financial-domain relationship extractor. Extract substantive mentions
of financial concepts by either participant in a conversation between an Investor
and a Financial Advisor.

For each substantive mention, extract two linked layers.

Layer A: evidence and provenance.

\texttt{raw\_subject}: the speaker who raised or discussed the entity; preserve
the transcript's speaker label.

\texttt{raw\_relation}: the surface-form relation phrase as spoken, when explicitly
expressed (e.g., ``interested in'' or ``asked about'').

\pagebreak
\texttt{raw\_object}: the exact words used in the transcript to refer to the entity.
Copy the surface form directly; do not paraphrase, abbreviate, expand, or infer.
For example, if the speaker says ``munis,'' write ``munis,'' not ``municipal bonds.''

\texttt{speaker\_role}: the role of the speaker, Investor or Financial Advisor.

\texttt{mention\_origin}: who raised the mention and in what conversational context,
e.g., ``investor proactive,'' ``investor responsive,'' or
``advisor\_introduced.''

\texttt{evidence\_text}: a verbatim quote containing the mention. Copy the exact
sentence(s) without modification, correction, or rephrasing.

\texttt{turn\_ids}: the turn(s) from which the evidence comes.

\texttt{confidence}: a value from 0.0 to 1.0.

Layer B: canonical information for graph construction.

\texttt{subject\_entity\_id}; \texttt{relation\_type}: \texttt{DISCUSSED}.

\texttt{object\_entity\_name}: the normalized/canonical entity name.

\texttt{object\_entity\_type}: the entity type, e.g., ``product'' or ``topic.''

Rules:
\begin{enumerate}
    \setlength{\itemsep}{0pt}\setlength{\parskip}{0pt}
    \item Extract only from the transcript. Do not infer facts from general knowledge.
    \item Copy \texttt{raw\_object}, \texttt{evidence\_text}, and any explicitly
    expressed \texttt{raw\_relation} verbatim. Preserve the transcript's speaker
    label in \texttt{raw\_subject}. The fields \texttt{speaker\_role},
    \texttt{mention\_origin}, \texttt{turn\_ids}, and \texttt{confidence} are
    annotations, not verbatim quotations.
    \item Match to the registry when possible in the canonical layer.
    Unknown entities must still be extracted.
    \item Return valid JSON only.
\end{enumerate}
\end{promptblock}

\noindent\textbf{Illustrative post-processing example.}
Suppose municipal bonds first appears in Turn~2, retirement income in Turn~5,
and municipal bonds is mentioned again in Turn~8. The resulting sequence is
municipal bonds followed by retirement income, yielding one directed transition
between them. The repeated mention supplies additional evidence but no newly
introduced concept. The LLM extracts mentions; post-processing constructs the flow.

\subsection{Transcript realization}\label{app:prompt-realization}

The realizer receives the sampled graph plan $G'$ and discourse plan $P$.
The graph plan uses the directed edges defined in Appendix~\ref{app:extended-methods}.
The slight-adjustment instruction below preserves flexible turn allocation;
it does not impose exact turn-by-turn execution.

\begin{promptblock}
\noindent\textbf{System prompt.}\par\nobreak
You are a dialogue writer specializing in realistic phone calls between a
Financial Advisor and an Investor.

Generate a synthetic transcript that:
\begin{enumerate}
    \setlength{\itemsep}{0pt}\setlength{\parskip}{0pt}
    \item Covers all nodes and edges from the provided graph plan.
    \item Follows the discourse plan: tone, goal, speaker pattern, and turn structure.
    \item Sounds like a natural spoken conversation. Use filler words,
    interruptions, and informal phrasing where appropriate.
    \item Does not invent key financial facts, product names, or entities beyond
    what is in the graph.
    \item Uses the turn plan as a guide, allowing slight adjustments
    (e.g., to turn allocation) for natural conversational flow.
\end{enumerate}

Output only the dialogue in this format:
\begin{samepage}
\begin{verbatim}
[Turn 1] Financial Advisor: ...
[Turn 2] Investor: ...
...
\end{verbatim}
\end{samepage}

\noindent\textbf{Input template.}\par\nobreak
\begin{samepage}
\begin{verbatim}
Graph plan G': {sampled_graph_plan}
Discourse plan P: {discourse_plan}
\end{verbatim}
\end{samepage}
\end{promptblock}

\subsection{Source-graph conditioning}\label{app:prompt-source-graph}

One training graph is supplied as the structural exemplar ($k=1$), without
source text. The exemplar is not a sampled DiRAFT graph plan. Call goal, tone,
speaker pattern, and turn budget are supplied alongside the exemplar, using
the same settings as the other corpus-conditioned strategies in the comparison.
Because realization instructions also differ, this comparison does not isolate
the effect of aggregate versus individual-graph conditioning.

\begin{promptblock}
\noindent\textbf{System prompt.}\par\nobreak
You are a dialogue writer specializing in realistic phone calls between a
Financial Advisor and an Investor.

You will be given one transcript-level knowledge graph extracted from a real
investor call. Generate one new conversation between an Investor and a Financial
Advisor, using this graph as a structural exemplar.

The transcript should:
\begin{enumerate}
    \setlength{\itemsep}{0pt}\setlength{\parskip}{0pt}
    \item Follow financial patterns similar to those in the example graph,
    including topics, products, and discourse flow.
    \item Follow the supplied call goal, tone, speaker pattern, and turn budget.
    \item Not copy or closely paraphrase the example; vary its patterns.
    \item Sound like a natural spoken conversation, using filler words,
    informal phrasing, and turn-taking.
    \item Cover a coherent set of financial topics and product interests,
    similar in scope to the example.
    \item Use realistic Financial Advisor language appropriate for an
    investor call.
\end{enumerate}

Output only the dialogue in this format, following the supplied turn budget:
\begin{samepage}
\begin{verbatim}
[Turn 1] Financial Advisor: ...
[Turn 2] Investor: ...
...
\end{verbatim}
\end{samepage}

\noindent\textbf{Input template.}\par\nobreak
\begin{samepage}
\begin{verbatim}
Source-graph exemplar:
{input_source_graph_exemplar}

Call goal: {call_goal}
Tone: {tone}
Speaker pattern: {speaker_pattern}
Turn budget: {turn_budget}
\end{verbatim}
\end{samepage}
\end{promptblock}

\newpage
\subsection{Scenario-only prompting}\label{app:prompt-scenario}

This baseline supplies a financial-call scenario and leaves concept order to the
generator (Appendix~\ref{sec:eval-realism}). It specifies 12--16 speaker turns
and approximately 500--700 words.

\begin{promptblock}
\noindent\textbf{Generation prompt.}\par\nobreak
You write fictional financial call transcripts for research.

Generate a plausible conversation between one Investor and one Financial Advisor
discussing investments.

Broad topics may include portfolio exposure, near-term liquidity needs, and
investment strategy. Choose the topic order naturally.

Use 12--16 speaker turns and approximately 500--700 words. Invent investor-specific
details and keep amounts, time horizons, and statements internally consistent.
Use generic market conditions rather than claims about current prices or events.

Output only the dialogue, with turns labeled ``Investor:'' and
``Financial Advisor:''.
\end{promptblock}

\subsection{Recoverability: Tier-3 semantic alignment}\label{app:prompt-alignment}

This Gemini~3.1 Pro pass matches residual entities after exact/substring and
lemmatized-token matching (Appendix~\ref{sec:eval-recoverability}). Unmatched
intended entities and still-available recovered entities are batched into one call
per sample. The response contains categorical pairings, not similarity scores.
This prompt specifies semantic alignment; it does not perform the separate
document-support assessment used by grounding-adjusted precision.

\begin{promptblock}
\noindent\textbf{System prompt.}\par\nobreak
You are an expert financial-advisory knowledge-graph analyst. You decide whether
two lists of concepts contain semantically equivalent entities. Two entities are
equivalent if a domain expert would treat them as referring to the same underlying
topic, product, or concept, even if the wording differs (synonyms, paraphrases,
abbreviations, or product/ticker equivalences). Do not match merely related-but-distinct
concepts.

\noindent\textbf{User prompt.}\par\nobreak
Intended entities (list A) that we could not match by string rules:
\begin{samepage}
\begin{verbatim}
A0: ...
A1: ...
...
\end{verbatim}
\end{samepage}

Recovered entities (list B) still available to match:
\begin{samepage}
\begin{verbatim}
B0: ...
B1: ...
...
\end{verbatim}
\end{samepage}

For each A entity, if there is a semantically equivalent B entity, output the
pairing. Each B may be used at most once. Return JSON of the form:
\begin{samepage}
\begin{verbatim}
{"matches": [
  {"a": <A index int>, "b": <B index int>},
  ...
]}
\end{verbatim}
\end{samepage}

Only include confident equivalences. Omit A entities with no equivalent.
\end{promptblock}

\newpage
\subsection{Blinded conversational-coherence judge}\label{app:prompt-coherence}

This rubric concerns the with/without-discourse-layer comparison in
Appendix~\ref{app:control-evaluation}. The value 3.5 is the interpretation boundary
between integer scores 3 and 4, not an additional response category or an
empirically calibrated threshold of human perception.

\begin{promptblock}
\noindent\textbf{System prompt.}\par\nobreak
You are a blinded expert judge evaluating the conversational coherence of calls
between a Financial Advisor and an Investor. You are shown two transcripts, A and
B. Score each transcript independently on a 1--5 scale.

Judge from the perspective of a human reader: how noticeable would coherence
problems be to an ordinary person reading the transcript? Scores below 3 denote
awkward transitions or breaks that a human would easily and obviously detect.
A score of 3 is borderline: minor rough spots a careful reader might notice but
that are not jarring and do not disrupt the overall flow. Scores of 4--5 denote
issues that are subtle or imperceptible, such that the conversation reads as
naturally human. Treat the boundary between ``human-noticeable'' and ``reads as
natural'' as lying between scores 3 and 4 (i.e., at 3.5). Reserve scores of 1--2
for problems that clearly and repeatedly stand out, and do not penalize a
transcript down to 3 or below for issues an ordinary reader would likely overlook.

Conversational Coherence (1--5):

1 = Fragmented and difficult to follow; a human would immediately find it incoherent.

2 = Frequent abrupt transitions or disconnected turns that a human would clearly
and repeatedly notice.

3 = Borderline: generally understandable with a few rough transitions a careful
reader might notice, but not jarring.

4 = Coherent and smooth, with only minor issues a human would barely notice.

5 = Highly coherent, with natural transitions and consistent conversational flow;
a human would perceive it as fully natural.

Consider logical progression, smooth topic transitions, turn-to-turn continuity,
and consistency of conversational purpose and tone. Combine these into one overall
coherence score per transcript; do not produce separate sub-metrics. Do not reward
verbosity or length.

\begin{samepage}
Respond with valid JSON only, exactly in this schema:

\begin{samepage}
\begin{verbatim}
{
  "A": {
    "coherence_score": <1-5 integer>,
    "reason": "One sentence"
  },
  "B": {
    "coherence_score": <1-5 integer>,
    "reason": "One sentence"
  }
}
\end{verbatim}
\end{samepage}
\end{samepage}

\noindent\textbf{User input template.}\par\nobreak
\begin{samepage}
\begin{verbatim}
Transcript A:
{transcript_A}

Transcript B:
{transcript_B}
\end{verbatim}
\end{samepage}
\end{promptblock}

\newpage
\subsection{Discourse-control identification judge}\label{app:prompt-control}

This author-approved standardized template describes the two-way setting
identification task judged by Gemini~3.1 Pro in
Appendix~\ref{app:control-evaluation}. Each invocation
specifies one control dimension and two candidate settings. Tone, call goal,
and speaker pattern share the assignment format; the template is distinct from
the coherence-rating task and human extractor validation.

\begin{promptblock}
\noindent\textbf{System prompt.}\par\nobreak
You are a blinded judge identifying discourse-control settings in conversations
between an Investor and a Financial Advisor.

You will receive:
\begin{enumerate}
    \setlength{\itemsep}{0pt}\setlength{\parskip}{0pt}
    \item A discourse-control dimension.
    \item Two candidate settings and their definitions.
    \item Two transcripts, A and B, with their assigned settings hidden.
\end{enumerate}

Evaluate only the specified dimension:
\begin{itemize}
    \setlength{\itemsep}{0pt}\setlength{\parskip}{0pt}
    \item Call goal: the main communicative purpose of the conversation.
    \item Tone: the communication style expressed in the conversation.
    \item Speaker pattern: which participant primarily directs the conversation,
    or whether both participants share that role.
\end{itemize}

Assign one candidate setting to each transcript. Use each setting exactly once.
Select the pairing best supported by the dialogue and the supplied definitions.
If the evidence overlaps, choose the better-supported overall pairing.

Base your decision on the specified control dimension. Do not let transcript
length or overall writing quality determine the assignment.

Return valid JSON only. Use the exact candidate setting names:
\begin{samepage}
\begin{verbatim}
{"A": "<setting name>", "B": "<setting name>"}
\end{verbatim}
\end{samepage}

\noindent\textbf{User prompt.}\par\nobreak
\begin{samepage}
\begin{verbatim}
Control dimension:
{control_dimension}

Candidate settings:
- {setting_1}: {definition_1}
- {setting_2}: {definition_2}

Transcript A:
{transcript_A}

Transcript B:
{transcript_B}
\end{verbatim}
\end{samepage}
\end{promptblock}

\noindent\textbf{Illustrative candidate settings.}
For tone, \emph{Exploratory} considers possibilities and asks open questions,
whereas \emph{Transactional} focuses on concrete actions and next steps.
For call goal, \emph{Portfolio review} reviews the investor's existing portfolio
and its alignment with investment objectives, whereas \emph{Specific inquiry}
addresses a particular investment question or concern.
For speaker pattern, \emph{Financial Advisor-led} means the Financial Advisor
primarily introduces topics and directs the discussion, whereas
\emph{Investor-led} assigns that role to the Investor.
These examples illustrate the input format rather than enumerate all tested pairs.

\end{document}